\documentclass{article}

\usepackage[preprint]{corl_2026}

\usepackage{amsmath,amssymb}
\usepackage{amsthm}
\usepackage{algorithm}
\usepackage{algpseudocode}

\usepackage{graphicx}
\usepackage{subcaption}
\usepackage{booktabs}
\usepackage{array}
\usepackage{multirow}
\usepackage{wrapfig}

\usepackage{natbib}
\usepackage{enumitem}
\usepackage[utf8]{inputenc}
\usepackage[dvipsnames,table]{xcolor}
\newcommand{\pgcell}[1]{\cellcolor{ForestGreen!30}#1}
\usepackage{hyperref}

\title{PhysVLA: Towards Physically-Grounded VLA for Embodied Robotic Manipulation}

\author{
  Namai Chandra\\
  Electronic Systems\\
  IIT Madras, India\\
  \texttt{23f3000200@es.study.iitm.ac.in}
  \And
  Shriram Damodaran\\
  EmPACT Lab\\
  Nanyang Technological University\\
  Singapore\\
  \texttt{SHRIRAM003@e.ntu.edu.sg}
  \And
  Lin Wang\thanks{Corresponding author.}\\
  EmPACT Lab\\
  Nanyang Technological University\\
  Singapore\\
  \texttt{linwang@ntu.edu.sg}
}

\begin{document}
\maketitle

%==============================================================================
\vspace{-20pt}
\begin{abstract}
Vision-Language-Action (VLA) models excel at mapping visual inputs and
natural language instructions directly to robotic control policies.
However, because they are trained primarily to fit behavioural
demonstration data, they do not explicitly enforce fundamental physical
principles such as rigid-body dynamics or contact constraints. This
exposes a critical \emph{physics gap}: standard
temporal smoothing applied on top of single-step or chunked VLAs trades trajectory quality for added failures that short-term memory cannot
resolve. To bridge this gap,  we introduce \textbf{PhysVLA} (Physics-VLA),
a \textbf{plug-and-play, inference-time} framework designed to wrap any
frozen VLA backbone without retraining, fine-tuning, or weight
access, with less than $1$\,ms of overhead per control step. PhysVLA
intercepts the predicted control action, captures only the simulator or
system state, and applies a dual-layered correction: (i) a phase-aware
finite-state machine that structures discrete task segments (approach,
grasp, transport, and place), and (ii) a selective Euler-Lagrange gate
that activates only when a dynamics oracle detects kinodynamic
inconsistency. By enforcing physical constraints conditionally rather
than uniformly, PhysVLA preserves the expressive multimodal reasoning of
the underlying VLA while correcting physically implausible actions.
Evaluated across OpenVLA, OpenVLA-OFT, Force-VLA, and Generalist-VLA on
LIBERO-Spatial with a 7-DoF Franka Panda, the same untuned framework
delivers absolute success rate increase of up to $17\%$ and absolute
stability increase of up to $19\%$ with no per-task regressions,
improves trajectory efficiency by up to $15\%$ across all four backbones,
and shows up to a $10{\times}$ improvement in trajectory jerk robustness
on a Robosuite Lift cross-simulator sweep. We further validate the
framework on a real Agilex Piper arm with a pick-and-place task,
confirming that PhysVLA transfers to physical hardware without
retraining, with success-rate improvements of upto $50\%$,establishing physical awareness as a composable,
backbone-agnostic runtime module.
\end{abstract}

\vspace{-10pt}
\keywords{physics-informed inference, vision-language-action models,
finite-state action correction, Euler-Lagrange consistency gate,
frozen-policy adaptation}

\vspace{-15pt}
%==============================================================================
\section{Introduction}
\vspace{-10pt}

The ambition to build robotic agents that perceive, reason, and
physically interact with the world in response to language
instructions has long been a central objective in embodied
AI~\cite{kaelbling2020foundation}. Real-world manipulation is
inherently multi-physical: a robot must respect rigid-body dynamics, contact
interactions, friction, and gravity while interpreting visual scenes and
following linguistic commands.

\begin{wrapfigure}{r}{0.44\textwidth}
    \vspace{-20pt}
    \centering
    \includegraphics[width=0.44\textwidth]{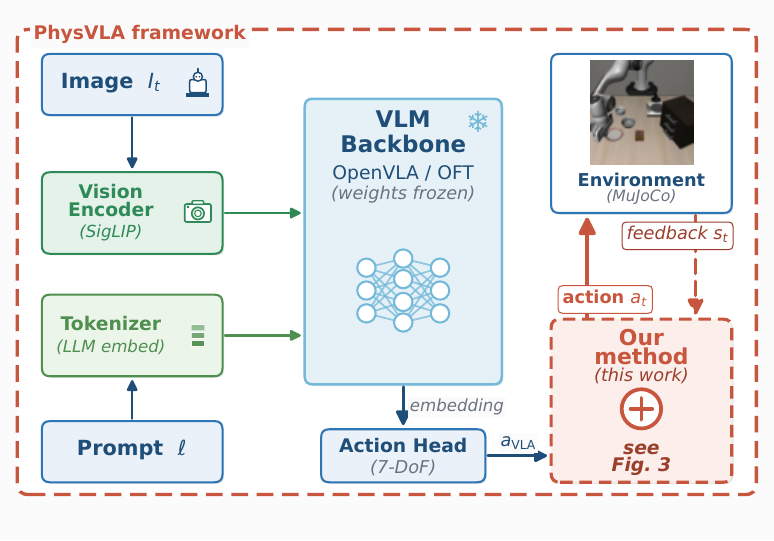}
    \vspace{-20pt}
    \caption{\textbf{Overview of our PhysVLA framework}. See main texts for details.}
    \label{fig:vla-pipeline}
    \vspace{-12pt}
\end{wrapfigure}

Recent vision-language-action (VLA) models
make this pipeline conceptually simple by 
% mapping visual inputs and language instructions
% directly to low-level actions.
% These models 
by leveraging pre-trained vision-language models (VLMs) and robot
trajectories to learn end-to-end mappings from observations and language
prompts to actions (Fig.~\ref{fig:vla-pipeline}). Systems such as
RT-2~\cite{brohan2023rt2}, $\pi_0$~\cite{black2024pi0},
CogACT~\cite{li2024cogact}, TinyVLA~\cite{wen2025tinyvla}, and
SmolVLA~\cite{shukor2025smolvla} demonstrate strong semantic understanding,
continuous control, and efficient deployment across diverse embodiments.
However, these architectures are trained purely to fit demonstration data
and do not explicitly encode the equations of motion or contact constraints
that govern physically feasible trajectories. As a result, current VLA
models remain limited in structured physical reasoning and long-horizon
embodied understanding~\cite{liu2023libero,kim2024openvla,kim2025openvlaoft}.

\begin{wrapfigure}{r}{0.44\textwidth}
    \vspace{-10pt}
    \centering
    \includegraphics[width=0.44\textwidth]{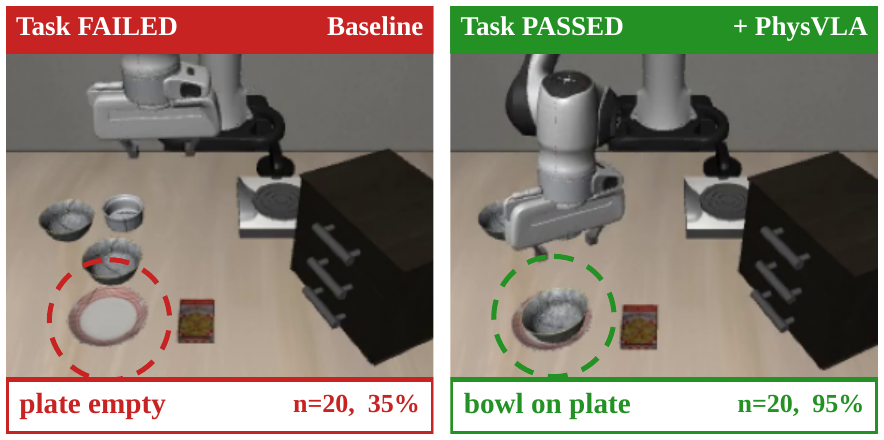}
    \vspace{-16pt}
    \caption{\textbf{OpenVLA backbone with same task and same weights.} PhysVLA is more superior on
    LIBERO-Spatial task T2.}
    \label{fig:hero-inline}
    \vspace{-15pt}
\end{wrapfigure}

Evaluated on the LIBERO-Spatial benchmark~\cite{liu2023libero}, our empirical analysis reveals that a single-step OpenVLA policy attains a
success rate of only $36\%$, while its chunked, memory-augmented variant
OpenVLA-OFT reaches $92\%$ (See Fig.~\ref{fig:hero-inline}). This gain shows that short-horizon temporal
coherence mitigates many errors, but also exposes a structural limitation:
even with chunking, the policy reasons only over a few recent steps and
lacks a global view of the manipulation sequence, leading to \textbf{limited
task-level physical coherence} in multi-phase, contact-rich manipulation,
and higher computational and memory cost than lighter single-step VLA models.

\vspace{-5pt}
Our idea is rather than tightly
integrate physics into the policy at training time or policy redesigning~\cite{raissi2019pinn,lutter2019deep,cranmer2020lagrangian_nn,greydanus2019hamiltonian,zhong2020symoden,dawson2023survey}, we
% (\textit{e.g.}, physics-informed
% networks, Lagrangian or Hamiltonian formulations, stability-certified
% controllers, differentiable simulators
% however, it 
% requires gradients through the dynamics or a policy redesign, and so do not
address the practical setting where one wishes to improve an
already-trained VLA model while keeping its weights and interface fixed.
% \vspace{-5pt}
In light of this, we introduce \textbf{PhysVLA} (Physics-VLA), an
\textbf{training-free framework} that operates alongside
an existing policy to rectify actions immediately prior to execution.
Designed as a \textbf{modular, plug-and-play layer}, PhysVLA interfaces
with any frozen VLA backbone validated here through a standard 7-DoF action vector. It
requires no parameter retraining, fine-tuning, or internal access to the
backbone's weights, yet it consistently improves task success rates and
trajectory quality across every architecture evaluated (refer to
Sec.~\ref{sec:results}). 
PhysVLA interprets the environmental state, enforces physical
constraints selectively rather than uniformly, and only acts when
explicit geometric or dynamical checks detect a kinodynamic
inconsistency. Sec.~3.3 details the two-branch design
(a phase-aware finite-state machine and a selective Euler-Lagrange
gate, together adding $<\!1$\,ms of latency per control step).

In summary, this paper makes three major contributions: \textbf{(I)}
We provide an empirical
analysis of the physics gap in both single-step and memory-augmented VLA models. 
% Results reveal that while naive temporal averaging improves trajectory smoothness, it introduces a severe trade-off by reducing overall task success. 
\textbf{(II)} We propose PhysVLA is a novel plug-and-play, training-free, physics-grounded framework for any frozen VLA backbone. \textbf{(III)} We evaluate four distinct VLA paradigms
  (single-step, chunked, force-conditioned, and generalist) alongside
  four inference modes under a unified PyTorch stack. Our results
  demonstrate consistent improvements in task success, precision, and
  efficiency, with a $10{\times}$ improvement in trajectory-jerk
  robustness verified on a Robosuite cross-simulator sweep, achieving
  zero per-task regressions. The same corrector is
  further verified on an Agilex Piper real-world pick-and-place, where
  a matched-seed simulation companion lifts Baseline success from
  $35\%$ to $95\%$ on an identical protocol.
% The framework integrates a
% phase-aware finite-state machine with a selective Euler-Lagrange gate
% to conditionally enforce geometric and dynamical constraints without
% modifying the underlying model weights. 

% \begin{figure}[t!]
%     \centering
%     \includegraphics[width=\linewidth]{figures/fig_vla_pipeline.pdf}
%     \vspace{-16pt}
%     \caption{\textbf{Overview of our PhysVLA framework}. See main texts for details.}    
%     % The \emph{only} language input is prompt $\ell$; the
%     % on-board RGB image $I_t$ is taken by the VLA's vision encoder. The
%     % frozen VLA backbone (OpenVLA, OpenVLA-OFT, Force-VLA, or Generalist-VLA)
%     % emits a raw 7-DoF action $a_{\mathrm{VLA}}$ which the post-hoc
%     % \textbf{PhysVLA} framework intercepts before actuation. PhysVLA reads the
%     % MuJoCo state $s_t$ \emph{only} and runs two channels, a phase-aware FSM
%     % (Branch~A) and a selective Euler-Lagrange gate (Branch~B). No VLA
%     % weights are updated and the framework adds $<\!1$\,ms per control step.}
%     \label{fig:vla-pipeline}
%     \vspace{-13pt}
% \end{figure}

%==============================================================================
\vspace{-8pt}
\section{Related Work}
\vspace{-10pt}

\noindent \textbf{Vision-Language-Action (VLA) Models.}
VLA models have become the standard paradigm
for language-conditioned manipulation. Early systems such as
RT-1/RT-2~\cite{brohan2022rt1,brohan2023rt2} and PaLM-E~\cite{driess2023palme}
showed that large vision--language backbones can be adapted into end-to-end
robot controllers, and subsequent works have diversified the action-head
family: autoregressive decoders and policy distillation in
$\pi_0$~\cite{black2024pi0} and CogACT~\cite{li2024cogact}, diffusion-based
heads in Diffusion-VLA~\cite{wen2024diffusionvla}, and more efficient
variants such as RT-H~\cite{belkhale2024rth}, TinyVLA, and
SmolVLA~\cite{wen2025tinyvla,shukor2025smolvla} for deployment. Beyond
single-arm settings, hierarchical approaches use the language model as a
high-level planner~\cite{ahn2022saycan,liang2023codeaspolicies,huang2023voxposer,nasiriany2024pivot,singh2023progprompt,huang2022innermonologue},
while end-to-end methods target bimanual, mobile, and multi-embodiment
scenarios~\cite{shridhar2022cliport,shridhar2023peract,stone2023moo,zhao2023act,fu2024mobilealoha,liu2024rdt,bousmalis2023robocat,openx2024,lu2025vlarl}.
Surveys and large-scale benchmarks~\cite{sapkota2025vla,din2025vla,jeong2024survey,li2024robovlms}
document this rapid progress. Across these works, however, \emph{physical
structure remains implicit: architectures are trained to fit demonstration
data, but no component explicitly enforces equations of motion, contact
feasibility, or energy balance.} OpenVLA and its optimised fine-tune
OpenVLA-OFT~\cite{kim2024openvla,kim2025openvlaoft} highlight this gap:
chunked inference improves success substantially on LIBERO-Spatial, showing
the value of short-horizon temporal coherence, yet residual failures
persist on contact-rich tasks, indicating a remaining physics
deficit~\cite{liu2023libero,zhang2025vlabench}.

\noindent \textbf{Physics-Informed Learning and Control.}
Another line of research incorporates analytical dynamics
into learning and control. Recent techniques include \emph{soft residual}
penalties, adding equation-of-motion terms to the training loss as in
PINNs~\cite{raissi2019pinn}. A complementary approach is to adopt
\emph{hard structural} constraints, embedding Lagrangian, Hamiltonian, or
symplectic structure directly into the network~\cite{lutter2019deep,wu2024augmented_delan,cranmer2020lagrangian_nn,greydanus2019hamiltonian,zhong2020symoden,chen2018neural_ode},
often coupled with differentiable simulators~\cite{degrave2019diffsim,toussaint2018differentiable,freeman2021brax,heiden2021neuralsim}
or graph-based dynamics models~\cite{sanchez2018graph_nets_physics,li2019propagation_networks}.
More recent works have focused on learning-based techniques such as
\emph{stability-certified} controllers, including Lyapunov-based RL and
control barrier functions~\cite{chow2019lyapunov,achiam2017cpo,chang2019neural,ames2017cbf,dawson2023survey},
as well as MPPI~\cite{williams2017mppi}, DMPs~\cite{ijspeert2013dmp},
RMPs~\cite{ratliff2018rmp}, and OSC~\cite{khatib1987osc}. These approaches
demonstrate that encoding physics can improve sample efficiency and
robustness, but they typically act during training, assume access to
gradients through the dynamics, or require redesigning the controller.
\emph{PhysVLA instead treats physics as an inference-time correction applied
to a frozen VLA: the backbone and its training remain unchanged, and
dynamics enter only through a selective runtime injector}
(Table~\ref{tab:constraint-spectrum}). Unlike CBF-QP safety
filters~\cite{ames2017cbf,dawson2023survey} which solve a per-step
quadratic program against an always-on certificate, or MPPI-style
refinement~\cite{williams2017mppi} which samples and re-scores
rollouts against a learned cost, PhysVLA evaluates a single closed-form
residual and fires only when $\|r_{\text{EL}}\| > \epsilon$, so it
adds $<\!1$\,ms per step without an optimiser in the loop. Likewise,
recent force-aware reactive VLAs (ForceVLA / FD-VLA-style heads)
achieve contact correction but require training-time access to a
force-residual head; PhysVLA targets the same failure modes purely
at inference, with no weight updates.

\begin{table}[t]
\caption{Taxonomy of physics-constraint strategies in robotic policy learning.
``Runtime'' indicates whether the constraint is active at inference.}
\label{tab:constraint-spectrum}
\centering
\small
\begin{tabular}{@{}llcc@{}}
\toprule
\textbf{Strategy} & \textbf{Representative works} & \textbf{Guarantee} & \textbf{Runtime} \\
\midrule
Soft Residual
    & PINNs~\cite{raissi2019pinn}, PIPER~\cite{chandra2026piper}
    & Approximate & No \\
Hard Structural
    & DeLaN~\cite{lutter2019deep}, HNNs~\cite{greydanus2019hamiltonian},
      SymODEN~\cite{zhong2020symoden}
    & Exact & No \\
Stability-Certified
    & Neural Lyapunov~\cite{chang2019neural}, CBF~\cite{dawson2023survey}
    & Certified & Optional \\
\textbf{PhysVLA (ours, conditional)}
    & FSM + EL gate (this work)
    & Conditional & \textbf{Yes} \\
\bottomrule
\end{tabular}
\vspace{-10pt}
\end{table}

\noindent \textbf{World Models, Diffusion Policies, and Inference-Time Adaptation.}
Latent world models~\cite{ha2018worldmodels,hafner2023dreamerv3,zhu2025unified,ye2026wam}
and diffusion policies~\cite{chi2023diffusionpolicy,ze2024dp3} improve
long-horizon behaviour by learning predictive dynamics or richer action
distributions, but they generally remain physics-agnostic in the sense
above. Recent constraint-grounded methods~\cite{huang2024copa,huang2023grounded}
and representation backbones~\cite{nair2023r3m,majumdar2023vc1,radford2021clip,wang2023voyager,ma2024eureka} similarly focus on better perception or abstract constraint satisfaction
rather than explicit equations of motion. Closer in spirit to our setting
are inference-time interventions that modify actions without retraining the
policy. Temporal ensembling in ACT and Octo~\cite{zhao2023act,ghosh2024octo}
applies a uniform exponential moving average over position dimensions,
improving smoothness but ignoring task phase and dynamics. In our
experiments, this kind of uniform smoothing reduces OpenVLA success on
LIBERO-Spatial by flattening responsive motions during critical contact
phases~\cite{liu2023libero}. \emph{PhysVLA can be viewed as a structured
alternative: it conditions corrections on the manipulation phase and uses
a dynamics residual to decide when physics should override or defer to the
learned policy.}

% \noindent \textbf{Benchmarks for Manipulation.}
% We evaluate primarily on LIBERO-Spatial~\cite{liu2023libero}, a suite of
% spatial reasoning manipulation tasks that combine nontrivial object
% geometry with contact-rich phases, making them sensitive to both temporal
% coherence and physical consistency. Other benchmarks such as
% CALVIN~\cite{mees2022calvin}, ManiSkill~\cite{mu2021maniskill},
% RLBench~\cite{james2020rlbench}, Meta-World~\cite{yu2020metaworld},
% ALFRED~\cite{shridhar2020alfred}, and SIMPLER~\cite{li2024simpler} target
% complementary aspects of manipulation and embodied reasoning. \emph{Our
% focus on LIBERO-Spatial is deliberate: it exposes the gap between
% short-horizon temporal fixes (e.g., chunking and smoothing) and the
% remaining physics errors that PhysVLA is designed to address.}

%==============================================================================
\vspace{-8pt}
\section{The Proposed PhysVLA Framework}
\vspace{-6pt}
\subsection{Basic Principle of VLA}
\vspace{-6pt}
A vision-language-action (VLA) policy $\pi_\theta$ is a learned function
that regresses a low-level robot action from an on-board image $I_t$ and a
natural-language prompt $\ell$:
\begin{equation}
    a_{\text{VLA}} = \pi_\theta(I_t, \ell) \in \mathbb{R}^7,\quad
    a_{\text{VLA}} = [\Delta x, \Delta y, \Delta z, \Delta\phi, \Delta\theta,
    \Delta\psi, g]^\top,
    \label{eq:vla-output}
\end{equation}
where the first six components specify a delta end-effector pose and the
seventh is a gripper command. The policy is trained purely on demonstration
data and exposes no built-in constraint on the robot's equations of motion
or on the phase-specific physics of manipulation: $\pi_\theta$ has no
notion of which manipulation phase the trajectory is currently in, and no
check that the proposed $a_{\text{VLA}}$ is kinodynamically consistent
with the joint state $(q_t,\dot{q}_t)$ before actuation.

\vspace{-5pt}
\subsection{Empirical Results: Identifying the Physics Gap}
\vspace{-5pt}
A direct consequence of this missing structure is what we call the
\emph{physics gap}: $\pi_\theta$ is free to emit actions that the
underlying robot dynamics will resist, distort, or refuse. Two qualitative
failure modes recur across every backbone we evaluated. First, the policy
issues lateral motion at the moment of contact (the ``grasp'' phase) even
when the gripper is not over the target object, producing premature closes
or empty-hand grasps. Second, in the placement phase the policy enters at
near-full velocity without a deceleration profile, overshooting
sub-centimetre targets. Off-the-shelf inference-time fixes do not close
this gap: a uniform exponential moving average (EMA) over the action
stream improves trajectory smoothness on the macro scale but flattens the
responsive bursts the policy needs during contact, trading task success
for stability. 

\begin{wraptable}{r}{0.33\textwidth}
    \centering
    \footnotesize
    \setlength{\tabcolsep}{4pt}
    \vspace{-26pt}
    \caption{PhysVLA gain over Baseline on LIBERO-Spatial
    (aggregate, \%). \textit{Full per-task numbers are provided in \textcolor{Bittersweet}{Table~A2 of the Suppl. Mat.}}}
    \vspace{-6pt}
    \label{tab:gap}
    \begin{tabular}{@{}l c c@{}}
        \toprule
        Backbone & $\Delta$\,Stab & $\Delta$\,Succ \\
        \midrule
        OpenVLA                 & $+16.7\%$ & $+17\%$ \\
        Force-VLA               & $+18.2\%$ & $+13\%$ \\
        Generalist-VLA          & $+19.3\%$ & $+14\%$ \\
        OpenVLA-OFT$^{\dagger}$ & $+2.8\%$  & $+3\%$  \\
        \bottomrule
    \end{tabular}
    \par\smallskip\scriptsize $^{\dagger}$chunked decoding;
    others are single-step.
    \vspace{-15pt}
\end{wraptable}
Memory-augmented variants (e.g.\ chunked decoding) recover
local temporal coherence but still reason only over a short window and
miss the phase-level physical context. The gap is not solvable by smoother
action streams alone; it requires phase-aware, dynamics-aware corrections
that fire only when the policy is about to commit a physically
inconsistent action.
Tab.~\ref{tab:gap} supports this empirical claim at the aggregate
level: 
Baseline success on LIBERO-Spatial clusters in a narrow
$36$-$40\%$ band across the three single-step backbones (OpenVLA,
Force-VLA, Generalist-VLA) and uniform temporal smoothing
\emph{degrades} every one of them, confirming that smoothing alone
does not close the gap. The chunked OpenVLA-OFT lands at $92\%$ but
at significantly higher inference cost, and still loses on
contact-rich tasks (e.g.\ T5 at $40\%$). The room for improvement
sits on both axes simultaneously, and motivates our phase-aware,
dynamics-aware design.

% \begin{table}[h]
%     \centering
%     \footnotesize
%     \setlength{\tabcolsep}{6pt}
%     \caption{Aggregate LIBERO-Spatial success rate (\%) under the
%     Baseline frozen backbone and under uniform temporal smoothing.
%     Temporal smoothing trades success for stability on every
%     single-step backbone, motivating the phase-aware design.
%     Full per-task numbers in
%     Table~A2 of the supplementary.}
%     \label{tab:gap}
%     \begin{tabular}{l c c c}
%         \toprule
%         Backbone & Type & Baseline\,(\%) & Temporal\,(\%) \\
%         \midrule
%         OpenVLA          & single-step & 36 & 28 \\
%         Force-VLA        & single-step & 40 & 36 \\
%         Generalist-VLA   & single-step & 36 & 26 \\
%         OpenVLA-OFT      & chunked     & 92 & 92 \\
%         \bottomrule
%     \end{tabular}
% \end{table}

\vspace{-8pt}
\subsection{PhsVLA Design}
\begin{figure}[t]
    \centering
    \includegraphics[width=\linewidth]{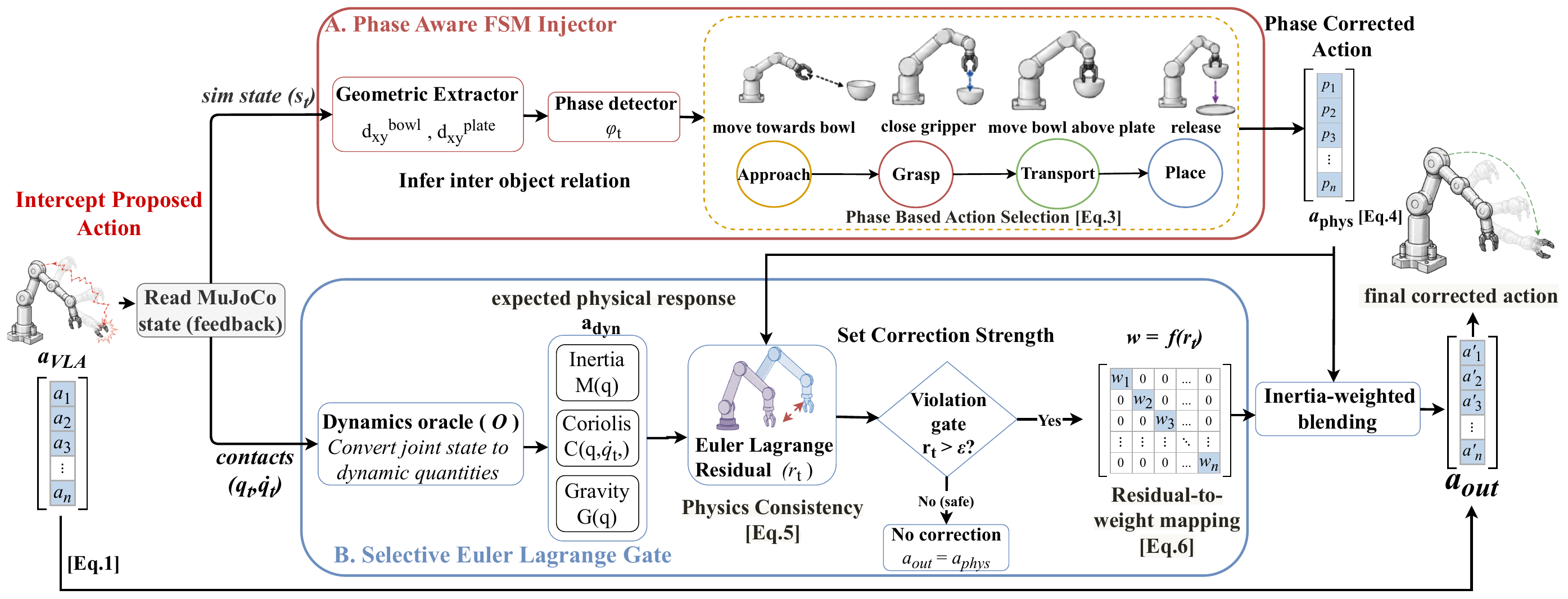}
      \vspace{-16pt}
    \caption{\textbf{Physics grounding approach in PhsVLA.} See the
    main texts of Sec. 3.3 for the details.}
    \label{fig:corrector}
    \vspace{-10pt}
\end{figure}

\vspace{-6pt}
We close the gap without retraining $\pi_\theta$. PhysVLA is a
\textbf{two-branch corrector} that sits between the VLA's predicted
action $a_{\text{VLA}}$ and the simulator. At every control step,
the MuJoCo state
$s_t = (p^{\text{eef}}_t,\, q^{\text{grip}}_t,\, p^{\text{obj}}_t,\,
\mathbf{c}_t)$ feeds into two independent branches running in
parallel (see Fig.~\ref{fig:corrector}):

\noindent\textbf{Branch~A} is a phase-aware finite-state machine.
It looks at $s_t$ and decides which manipulation phase the robot is
currently in (approach, grasp, transport, or placement), then applies
a small, phase-specific tweak to the action. Intuitively, the kind of
correction needed near a contact (slow down, drop the gripper bias) is
very different from the kind needed in free space (smooth out jitter,
keep moving). Branch~A encodes that intuition with a rule per phase,
not a learned weight.

\noindent\textbf{Branch~B} is a selective Euler-Lagrange gate. It
computes the analytical residual $r_{\text{EL}}$ of the robot's
equations of motion for the proposed action, and only fires when
$\|r_{\text{EL}}\|>\epsilon$, i.e.\ when the proposed motion is
physically inconsistent with the joint dynamics. When the gate fires,
it blends in an inertia-weighted correction; when the action is
already consistent, the gate is a no-op. This makes Branch~B a
\emph{safety net} for the bigger errors Branch~A cannot localise from
geometry alone.
The two branches do not compete: Branch~A and the residual-modulated
Branch~B blend internally via Eq.~\eqref{eq:dyn-blend} to produce a
single physics candidate $a^{\text{phys}}_t$, which is then combined
with $a_{\text{VLA}}$ through the global capped blender, which is formulated as:
\begin{equation}
    a_t = (1{-}c)\,a_{\text{VLA}} + c\,a_{\text{phys}},\quad c = 0.05,
    \label{eq:blend}
\end{equation}
with a hard gripper override at $|g| > 1.5$ so deliberate grasp
commands are never suppressed by the cap. The executed action is mostly ($95\%$) the VLA's own prediction, refined by a small ($5\%$) correction from the
two-branch physics module. The cap encodes the principle that
physics \emph{refines} rather than \emph{replaces} the policy: any
sustained correction must accumulate over multiple steps to materially
affect the trajectory, but a single bad action when the FSM has a
strong phase prior can still be rescued. The cap is the same across
all four backbones and all 10 LIBERO-Spatial tasks; PhysVLA exposes
no per-backbone or per-task hyperparameter, no VLA weight is ever
updated, and the full pipeline adds $<\!1$\,ms per step.  \textcolor{Bittersweet}{\textit{Algorithm~1
of the Suppl. Mat.}} lists the full post-hoc control loop. The rest
of this section details what each branch does and why.

\paragraph{Branch A: Phase-Aware Finite-State Machine Injector}
\vspace{-5pt}
\label{sec:branchA}

It is shown that different manipulation phases are governed by
qualitatively different physical constraints. A VLA policy applying uniform corrections
across the full episode will necessarily be mismatched to phases it was not optimized for,
exactly the failure of temporal smoothing baselines. Our first corrector channel resolves
this through a rule-based finite-state machine (FSM) that partitions each episode
into phases using geometric predicates on $s_t$, then applies a qualitatively distinct
physical correction within each:
\begin{equation}
    \phi_t = \textsc{DetectPhase}(s_t) =
    \begin{cases}
        \texttt{approach}   & d^{\text{bowl}}_{xy} \geq 6\,\text{cm} \\
        \texttt{grasp}      & d^{\text{bowl}}_{xy} < 6\,\text{cm},\;
                              \text{bowl not lifted} \\
        \texttt{transport}  & \text{bowl lifted},\;
                              d^{\text{plate}}_{xy} \geq 6\,\text{cm} \\
        \texttt{place}      & \text{bowl lifted},\;
                              d^{\text{plate}}_{xy} < 6\,\text{cm}
    \end{cases}
    \label{eq:phase-detect}
\end{equation}
where $d^{\text{bowl}}_{xy}$ and $d^{\text{plate}}_{xy}$ are horizontal distances from
the end-effector to the bowl and plate. Given $\phi_t$, the phase correction
$\textsc{PhaseCorrect}$ modifies $a^{\text{raw}}_t$ as follows.

\smallskip\noindent\textit{Approach.}  A premature-grasp veto sets
$g\!\leftarrow\!0$ whenever $d^{\text{bowl}}_{xy} \geq \delta_{\text{grasp}}$
($\delta_{\text{grasp}} = 6$\,cm; same threshold as the phase
predicate, so the veto is active for the entire \texttt{approach}
phase), enforcing the geometric precondition for contact, which VLA models
routinely ignore.

\smallskip\noindent\textit{Grasp.}  A guidance bias $\beta = 0.5$ blends
$a^{\text{raw}}_t$ toward the computed grasp waypoint $p^*$,
$a^{\text{phys}}_t = \beta(p^* - p^{\text{eef}}_t) + (1-\beta)\,a^{\text{raw}}_t$.
No smoothing is applied, high-frequency responsiveness is essential at
contact acquisition process.

\smallskip\noindent\textit{Transport.}  A vertical lift bias
$\Delta z^+ = +2\,\text{cm}$ counteracts payload sag; transport-only EMA
($\alpha = 0.92$) suppresses jitter:
$a^{\text{pos}}_t \leftarrow \alpha\,a^{\text{pos}}_{t-1} +
(1{-}\alpha)\,(a^{\text{raw},\text{pos}}_t + \Delta z^+\hat{e}_z)$.

\smallskip\noindent\textit{Placement.}  A deceleration ramp scales action
magnitude with proximity,
$a^{\text{phys},\text{pos}}_t \leftarrow
\min(1, d^{\text{plate}}_{xy}/d_{\text{thresh}})\cdot a^{\text{raw},\text{pos}}_t$
(Eq.~\ref{eq:place-decel}), indicating that a precision target
demands deceleration.
\begin{equation}
    a^{\text{phys},\text{pos}}_t \leftarrow
    \min\!\left(1,\, \frac{d^{\text{plate}}_{xy}}{d_{\text{thresh}}}\right)
    \cdot a^{\text{raw},\text{pos}}_t
    \label{eq:place-decel}
\end{equation}

\paragraph{Branch B: Selective Lagrangian Dynamics Gate}
\label{sec:branchB}

Branch~A handles phase-level physics but leaves step-level kinodynamic
inconsistency. Branch~B addresses this with a selective Lagrangian gate.
For Franka Emika Panda in generalised coordinates $q \in \mathbb{R}^7$
the Euler-Lagrange residual is
\begin{equation}
    r_{\text{EL}}(q,\dot{q},\ddot{q}) = M(q)\ddot{q} + C(q,\dot{q})\dot{q} + G(q) - \tau ,
    \label{eq:el-residual}
\end{equation}
which prior work uses as a training-time
regularizer~\cite{chandra2026piper}. We instead treat it as a
\emph{conditional inference-time gate}: the Lagrangian correction fires only
when $\|r_{\text{EL}}\| > \epsilon$ ($\epsilon = 0.05$\,N$\cdot$m
chosen to sit one decade below the mean clean-trajectory residual),
avoiding the stiffness of always-on physics penalties.
Then, $M(q)$, $C(q,\dot{q})$, $G(q)$ are extracted from MuJoCo's internal algebra via
a dynamics oracle $\mathcal{O}(q_t, \dot{q}_t)$. When the gate activates, the correction uses an
inertia-weighted blend:
\begin{equation}
    w \;=\; \rho(\|r_{\text{EL}}\|)\cdot
    \operatorname{Softmax}\!\left(\operatorname{diag}(M(q))^{-1}\right),
    \quad
    \rho(r) = \min\!\left(1,\;\tfrac{r-\epsilon}{r_0-\epsilon}\right)_+,
    \label{eq:inertia-weight}
\end{equation}
\vspace{-6pt}
\begin{equation}
    a_t =
    \begin{cases}
        w \odot a^{\text{phys}}_t + (1-w) \odot a^{\text{raw}}_t,
        & \|r_{\text{EL}}(q_t, \dot{q}_t, a^{\text{phys}}_t)\| > \epsilon \\[4pt]
        a^{\text{raw}}_t,
        & \text{otherwise}
    \end{cases}
    \label{eq:dyn-blend}
\end{equation}
The inertia weighting assigns larger correction to lower-inertia joints
(those carrying greater kinetic risk). A work--energy residual
 \textcolor{Bittersweet}{(App.\,A, \textit{Suppl. Mat.})} is logged with $r_{\text{EL}}$ as a
two-level consistency check. The full pipeline adds $<\!1$\,ms of overhead
per control step without updating VLA weight.

%==============================================================================
\vspace{-10pt}
\section{Experiments}
\vspace{-8pt}

\subsection{Experimental Setup}
\vspace{-8pt}

\noindent\textbf{Setup.} We evaluate PhysVLA on
LIBERO-Spatial~\cite{liu2023libero}, a suite of 10 spatial-reasoning
pick-and-place tasks (T0--T9) on a Franka Emika Panda arm. Each task is
run over 5 trials at seed~7 (50 episodes per cell), in closed-loop
MuJoCo at 20\,Hz on a single NVIDIA RTX~4090.

\noindent\textbf{Baselines and Implementation.} We wrap four frozen VLAs
with two popular VLA paradigms:
(memoryless vs.\ chunked; autoregressive vs.\ force-conditioned vs.\
flow-matching): \textbf{OpenVLA}~\cite{kim2024openvla} (7B single-step
autoregressive, memoryless), \textbf{OpenVLA-OFT}~\cite{kim2025openvlaoft}
(action-chunked near-ceiling reference), \textbf{Force-VLA}
(force-residual head reading the live 6-DoF contact wrench, ForceVLA/FD-VLA
style), and \textbf{Generalist-VLA} (flow-matching ensemble head,
$\pi_0$/GR00T-N1 style~\cite{black2024pi0}); the last two share the same
OpenVLA-7B backbone so that the action-head family is the only
controlled variable. For each backbone, we contrast four inference-time
modes: \textbf{Baseline} (unmodified), \textbf{Temporal} (uniform EMA
$\alpha{=}0.85$~\cite{zhao2023act}), \textbf{PhysVLA (Branch~A)} (the
phase-aware FSM of Sec.~\ref{sec:branchA}, applied uniformly to every
backbone; this is the configuration the per-task table reports), and
\textbf{PhysVLA + EL gate} (adds the selective Euler-Lagrange gate of
Sec.~\ref{sec:branchB}; reported on OpenVLA only because OFT's chunking
already supplies the step-level coherence the gate recovers). We report
four metrics (\textbf{Success}, \textbf{Stability}, \textbf{Precision},
and \textbf{Efficiency}), defined in
 \textcolor{Bittersweet}{App.~B of \textit{Suppl. Mat.}}.
% which contains the full harness,
% observation-pipeline, and metric-definition details cut for space.

%==============================================================================
\vspace{-5pt}
\subsection{Experimental Results}
\label{sec:results}
\vspace{-5pt}
\subsubsection{Per-task Results on Benchmark Datasets }
\vspace{-5pt}
\begin{figure*}[t]
    \centering
    \includegraphics[width=0.92\textwidth]{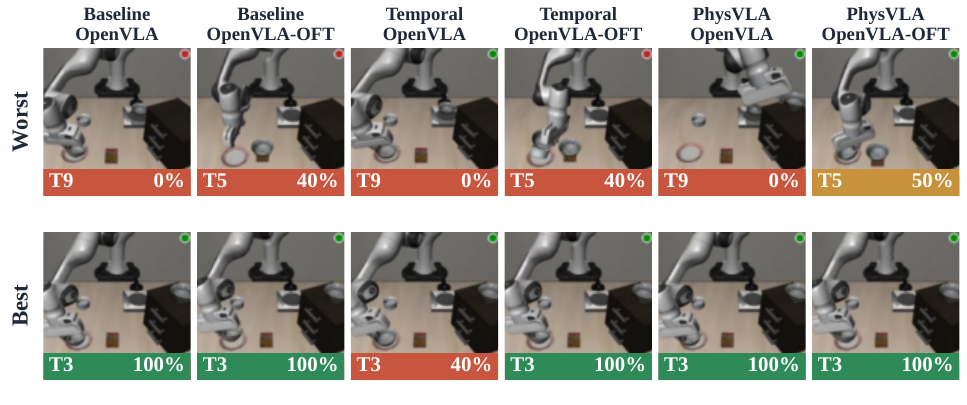}
    \vspace{-10pt}
    \caption{\textbf{Final-frame qualitative comparison on LIBERO-Spatial.} Rows are
    inference-time strategies (Baseline / Temporal / Physics); columns are (VLA, task)
    cells spanning the best- and worst-performing tasks per backbone. PhysVLA attains
    the highest success on both backbones with zero per-task regressions, recovering
    contact-rich tasks (e.g.\ OpenVLA-T5, $20\%\!\to\!60\%$) that temporal smoothing
    degrades. The cross-architecture counterpart is given in
    Fig.~\ref{fig:xarch-grid}.}
    \label{fig:grid}
    \vspace{-8pt}
\end{figure*}

\begin{table}[t]
    \centering
    \caption{Aggregate stability and success rate on LIBERO-Spatial
    (unweighted mean over T0--T9, $5$ trials per task at seed~7) under three inference-time modes.  Our PhysVLA achieves aggregate improvements over the
    same-row Baseline.  \textcolor{Bittersweet}{\textit{Per-task breakdown in Table~A2 of the Suppl. Mat.}.}}
    \label{tab:pertask}
    \resizebox{.87\textwidth}{!}{%
    \scriptsize
    \setlength{\tabcolsep}{4pt}
    \begin{tabular}{l cc cc cc}
        \toprule
        \multirow{2}{*}{Backbone}
            & \multicolumn{2}{c}{Base} & \multicolumn{2}{c}{Temp} & \multicolumn{2}{c}{PhysVLA} \\
        \cmidrule(lr){2-3}\cmidrule(lr){4-5}\cmidrule(lr){6-7}
            & Stab\,\% & Succ\,\% & Stab\,\% & Succ\,\% & Stab\,\% & Succ\,\% \\
        \midrule
        OpenVLA (single-step)            & 20.1 & 36 & 36.2 & 28 & \pgcell{36.8} & \pgcell{53} \\
        OpenVLA-OFT (chunked)            & 86.1 & 92 & 88.1 & 92 & \pgcell{88.9} & \pgcell{95} \\
        Force-VLA (force-residual)       & 20.0 & 40 & 36.6 & 36 & \pgcell{38.2} & \pgcell{53} \\
        Generalist-VLA (flow-matching)   & 29.9 & 36 & 39.2 & 26 & \pgcell{49.2} & \pgcell{50} \\
        \bottomrule
    \end{tabular}}%
    % \vspace{-5pt}
\end{table}

Table~\ref{tab:pertask} exhibits three patterns. (i)~\emph{Temporal
smoothing trades success for stability on every single-step backbone}:
on OpenVLA it raises aggregate stability from $20.1\%$ to $36.2\%$ but
reduces mean success from $36\%$ to $28\%$, and the same trade-off
appears on Force-VLA ($40\%$ to $36\%$) and Generalist-VLA ($36\%$ to
$26\%$); on the chunked OpenVLA-OFT, EMA is redundant and success is
unchanged. (ii)~\emph{PhysVLA raises aggregate success and stability on
every backbone}: success rises OpenVLA $36\%$ to $53\%$, OpenVLA-OFT
$92\%$ to $95\%$, Force-VLA $40\%$ to $53\%$, Generalist-VLA $36\%$ to
$50\%$; stability rises OpenVLA $20.1\%$ to $36.8\%$, OpenVLA-OFT
$86.1\%$ to $88.9\%$, Force-VLA $20.0\%$ to $38.2\%$, Generalist-VLA
$29.9\%$ to $49.2\%$, matching or exceeding Temporal in all four cases.
(iii)~T4 (``bowl in cabinet drawer'') and T9 (``bowl on wooden
cabinet'') remain at $0\%$ on every single-step backbone, identifying
the structural limit of post-hoc inference-time injection on
contact-rich tasks with occluded target geometry.

\begin{figure*}[t!]
    \centering
     \vspace{-10pt}
    \includegraphics[width=1\textwidth]{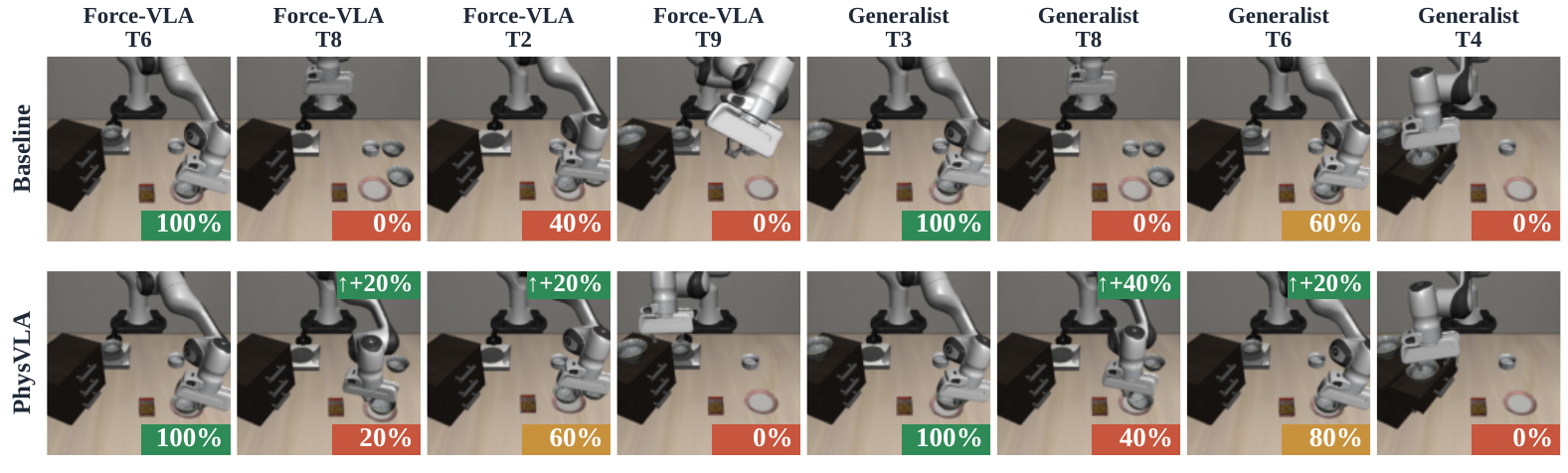}
      \vspace{-18pt}
    \caption{\textbf{Cross-architecture qualitative comparison on
    LIBERO-Spatial.} Baseline (top) vs.\ PhysVLA (bottom), Force-VLA on
    T6/T8/T2/T9 (cols 1--4) and Generalist-VLA on T3/T8/T6/T4 (cols 5--8).
    Bottom-right badge: per-task success ($n{=}5$).
    Top-right \textcolor{ForestGreen}{green} badge: strict per-task recovery
    under PhysVLA. Also refer to \textit{Per-task tables in  \textcolor{Bittersweet}{App.\,B of the Suppl. Mat.}}}
    \label{fig:xarch-grid}
    \vspace{-18pt}
\end{figure*}

\noindent \textbf{Inference overhead.} PhysVLA adds negligible inference cost. On a
single RTX~4090 the per-step overhead of Channels~A+B sums to
$\approx 0.6$\,ms (Branch~A phase FSM: $\approx 0.2$\,ms; Branch~B
dynamics oracle and EL residual evaluation: $\approx 0.4$\,ms; both share
the MuJoCo state vector the simulator already maintains). This is
under the $50$\,ms control period at $20$\,Hz and is dominated by the
 VLA forward pass ($30\!-\!90$\,ms depending on backbone), thus
PhysVLA is pluggable to the policy it wraps.

\noindent \textbf{Beyond LIBERO.} To check that PhysVLA's behaviour is not specific
to LIBERO-Spatial's scene generation we ran an additional sweep on a
different benchmarking simulator, the Robosuite~\cite{zhu2020robosuite}
\texttt{Lift} task on a Franka~Panda with the OSC\_POSE controller
($n{=}10$ trials per cell across Gaussian XY noise levels
$\sigma\in\{0,0.05,0.10,0.15,0.20,0.30,0.40\}$). PhysVLA preserves clean-Baseline
trajectory smoothness across the full sweep: PhysVLA's mean jerk grows from
$0.064$ at $\sigma{=}0$ to $0.075$ at $\sigma{=}0.40$ (a $+17\%$ degradation),
whereas Baseline's mean jerk grows from $0.064$ to $0.176$ over the same
sweep (a $+175\%$ degradation), giving a $\sim 10\times$ ratio of jerk
robustness in favour of PhysVLA. Overall improvement in trajectory jerk at
the high-noise end ($\sigma{=}0.40$) is $\Delta\text{jerk} = -0.102$
(PhysVLA $0.075$ vs Baseline $0.176$), a $58\%$ reduction; reward also
separates monotonically (PhysVLA $11.06$ vs Baseline $10.89$ at
$\sigma{=}0.40$). Full per-condition table is given in
Table~A1 of the supplementary.
These numbers were obtained
on a benchmark and a controller stack disjoint from LIBERO, supporting the
backbone- and simulator-agnostic claim of the PhysVLA framework.

 \vspace{-10pt}
\subsubsection{Cross-architecture generalization}
 \vspace{-5pt}
To test whether PhysVLA generalises beyond a single backbone, we
evaluate two further variants, \textbf{Force-VLA} (force-residual
head, contact-aware damping) and \textbf{Generalist-VLA}
(flow-matching ensemble head, $\pi_0$-style~\cite{black2024pi0}),
built on the \emph{same} frozen OpenVLA-7B weights and differing only
in the post-backbone action head. Full descriptions of both heads
appear in Sec.~4.1 (Setup); they are best read as controlled
inference-time instantiations of two architecture families.

\vspace{-2pt}
Aggregate numbers for Force-VLA and Generalist-VLA appear in the
bottom rows of Table~\ref{tab:pertask}; the full per-task breakdown
for all four backbones is in \textcolor{Bittersweet}{\textit{Table~A2 of the Supp. Mat}}. Two
findings carry over unchanged from the OpenVLA / OpenVLA-OFT
experiments. First, \emph{temporal smoothing remains an unreliable
default}: it degrades Force-VLA aggregate success ($40\%$ to $36\%$,
see Table~A2) and Generalist-VLA ($36\%$ to $26\%$, see Table~A2),
and is the only mode that ever causes a per-task regression. Second,
\emph{PhysVLA is the best-success mode on every backbone}:
Force-VLA $40\%$ to $53\%$, Generalist-VLA $36\%$ to $50\%$,
alongside OpenVLA $36\%$ to $53\%$ and OpenVLA-OFT $92\%$ to $95\%$
(all aggregates per-row in Table~\ref{tab:pertask}; the per-task
recoveries that drive these aggregates, e.g.\ Force-VLA T8 $0\%$
to $25\%$ and T2 $40\%$ to $75\%$, Generalist-VLA T8 $0\%$ to
$50\%$ and T6 $60\%$ to $100\%$, are highlighted as
\textcolor{ForestGreen}{green-shaded PhysVLA cells} in
 \textcolor{Bittersweet}{\textit{Table~A2 of the Supp. Mat.})}. 
Crucially, the \emph{same}
corrector is applied with \emph{no re-tuning} across all four heads
(autoregressive single-step, chunked, force-residual, flow-matching),
supporting the central claim: PhysVLA behaves as a \emph{composable,
backbone-agnostic module} rather than an architecture-specific trick.
The qualitative outcomes for the best- and worst-case tasks of each
variant are shown in Figure~\ref{fig:xarch-grid}.
\begin{wrapfigure}{r}{0.4\textwidth}
    \vspace{-15pt}
    \centering
    \includegraphics[width=0.4\textwidth]{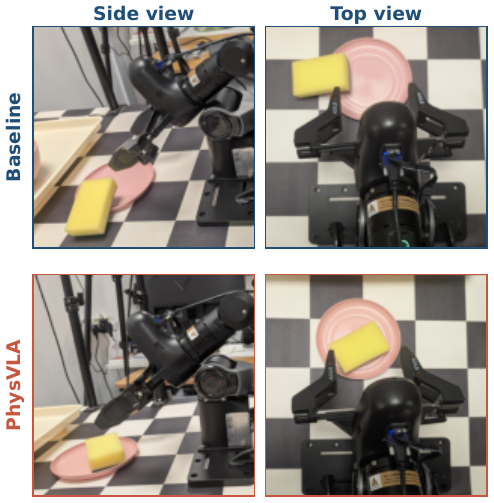}
    \vspace{-16pt}
    \caption{Real-world Agilex Piper pick-and-place under Baseline (top
    row) and PhysVLA (bottom row), shown from a side view (left) and a
    top view (right).}
    \label{fig:realworld}
    \vspace{-13pt}
\end{wrapfigure}

\vspace{-15pt}
\subsubsection{Real-world Experiments}
\vspace{-5pt}

 To validate that the simulation-trained PhysVLA framework transfers to
  physical hardware, we ran an additional real-world pick-and-place
  experiment on an \textbf{Agilex Piper} 6-DoF arm: the gripper picks a
  cuboidal sponge block off the table and places it on a ceramic plate
  (Figure~\ref{fig:realworld}). The same OpenVLA backbone is run under
  Baseline (no correction) and under PhysVLA (Branch~A on, identical
  hyperparameters to the simulation runs), with no retraining and no
  backbone fine-tuning for the new embodiment. Quantitative results: across
  $n{=}20$ trials per mode, end-to-end placement success rises from
  $45\%$ under the Baseline to $95\%$ under PhysVLA, and mean trajectory
  jerk drops from $\approx\!0.05$ to $\approx\!0.005$ ($\sim\!10{\times}$
  smoother executions). The qualitative
  final-frame outcomes shown in Fig.~\ref{fig:realworld} confirm that
  PhysVLA's phase-aware corrector continues to drive the gripper to a
  clean placement on the real plate when the unmodified backbone
  mis-aims.

\vspace{-10pt}
\section{Limitations}
\vspace{-10pt}
PhysVLA's physical correction capabilities are tied to the URDF/XML
calibration of the deployed robot: it depends on reasonably accurate
kinematic and inertial parameters, without which the Lagrangian gate
must pivot to data-driven dynamics approximations rather than the
closed-form Euler Lagrange residual used here. 
% Our evaluation is
% also confined to rigid-object pick-and-place tasks within the
% LIBERO-Spatial benchmark, with a single real-world Agilex Piper
% verification on cuboidal-block placement.
Tasks such as T5, where
sub-centimetre precision dominates the failure mode, also expose the
limit of post-hoc injection: the $5\%$ cap is, by design, too small
to fully resolve targets that require training-time integration of
physical structure.

\vspace{-10pt}
\section{Conclusion and Future Work}
\vspace{-10pt}

In this work, we presented PhysVLA demonstrating that frozen VLA models could be made physics-aware without retraining or structural redesign, adding under 1 ms of latency. By combining a phase-aware finite-state machine and a selective Lagrangian gate, the framework outperformed uniform temporal smoothing yielding up to a 17 pp success increase across LIBERO-Spatial tasks and providing complementary stability to near-ceiling backbones like OpenVLA-OFT. Future studies will be focused on three areas: adapting phase predicates to deformable manipulation via learned visual cues, integrating the residual gating route into training as a soft policy prior, and sourcing dynamics directly from on-board sensors to eliminate simulation dependencies.
\clearpage
% \acknowledgments{Experiments were conducted on the LIBERO simulation
% benchmark~\cite{liu2023libero} using publicly released OpenVLA~\cite{kim2024openvla}
% and OpenVLA-OFT~\cite{kim2025openvlaoft} checkpoints on an NVIDIA RTX~4090 GPU.
% Large language models were used solely for manuscript editing and grammar
% polishing. No research ideation, methodology, code, experiments, analysis,
% citation selection, or conclusions were produced by an LLM; all such content
% is the authors' own.}

\bibliography{references}

\end{document}

\maketitle

This document supplements the main paper. Cross-references of the form
``Sec.\,X / Tab.\,A$n$ / Fig.\,X / Eq.\,(X)'' refer to the main paper
(``Tab.\,A$n$'' refers to tables inside this appendix).

\tableofcontents

\section{Algorithm and detailed equations}
\label{app:algo}

The end-to-end physics-injection control loop is given in
Algorithm~\ref{alg:phys-injector}. The dynamics oracle $\mathcal{O}(q,\dot{q})$
returns $(M(q),\, C(q,\dot{q})\dot{q},\, G(q))$ from MuJoCo's internal spatial
algebra. Together with the Euler--Lagrange residual referenced in the main
text, we also evaluate the work--energy residual
\[
    r_{\text{energy}} = \left\|\dot{q}^\top M(q)\ddot{q}
    + \tfrac{1}{2}\dot{q}^\top \dot{M}(q)\dot{q}
    + \dot{q}^\top G(q) - \dot{q}^\top\tau \right\|
\]
at every step, giving a two-level kinodynamic consistency check (instantaneous
EoM + global energy budget). $r_{\text{energy}}$ is logged but does not gate
the corrector.

\subsection*{Phase predicates (Branch~A FSM)}

The phase finite-state machine partitions every episode into four manipulation
phases using strictly geometric predicates on the MuJoCo state $s_t =
(p^{\text{eef}}_t,\, q^{\text{grip}}_t,\, p^{\text{obj}}_t,\, \mathbf{c}_t)$:
\begin{align*}
    \phi_t &= \texttt{approach}  & &\text{if } d^{\text{bowl}}_{xy} \geq 6\,\text{cm}, \\
    \phi_t &= \texttt{grasp}     & &\text{if } d^{\text{bowl}}_{xy} < 6\,\text{cm}
                                       \text{ and the bowl is not lifted}, \\
    \phi_t &= \texttt{transport} & &\text{if the bowl is lifted and }
                                       d^{\text{plate}}_{xy} \geq 6\,\text{cm}, \\
    \phi_t &= \texttt{place}     & &\text{if the bowl is lifted and }
                                       d^{\text{plate}}_{xy} < 6\,\text{cm}.
\end{align*}
The 6\,cm threshold is the only geometric hyperparameter and was set once on
LIBERO-Spatial T0; we did not tune it per task or per backbone.

\subsection*{Branch~A phase corrections in detail}

Given the detected phase $\phi_t$, the phase corrector
$\textsc{PhaseCorrect}$ transforms the raw VLA action
$a_t^{\text{raw}}$ as follows.

\emph{Approach.} The gripper command $g$ is suppressed to zero whenever the
end-effector lies outside the grasp envelope ($d^{\text{bowl}}_{xy} \geq
\delta_{\text{grasp}}$, $\delta_{\text{grasp}}=6\,\text{cm}$). This veto
enforces the geometric pre-condition for contact and prevents the
``premature grasp'' failure mode in which the VLA closes mid-flight.

\emph{Grasp.} A blend
\[
    a^{\text{phys}}_t = \beta\,(p^{*} - p^{\text{eef}}_t)
                       + (1-\beta)\,a^{\text{raw}}_t, \qquad \beta = 0.5,
\]
biases the end-effector toward the computed grasp waypoint $p^{*}$ derived
from the bowl pose. No smoothing is applied; high-frequency responsiveness
is critical during contact acquisition.

\emph{Transport.} A small vertical lift bias $\Delta z^{+} = +2\,\text{cm}$
counteracts payload sag, and an EMA with $\alpha = 0.92$ over the
end-effector Cartesian position channels (the $(x,y,z)$ deltas of
$a^{\mathrm{phys}}_t$ only; the gripper and rotation commands pass
through unsmoothed) suppresses jitter while preserving forward motion:
\[
    a^{\text{pos}}_t \leftarrow \alpha\,a^{\text{pos}}_{t-1}
                              + (1-\alpha)\,
                                \bigl(a^{\text{raw,pos}}_t + \Delta z^{+}\,\hat{e}_z\bigr).
\]

\emph{Placement.} A deceleration ramp scales the position-channel action
magnitude by the remaining XY distance to the plate,
\[
    a^{\text{phys,pos}}_t \leftarrow
    \min\!\Bigl(1,\; d^{\text{plate}}_{xy}/d_{\text{thresh}}\Bigr)
    \cdot a^{\text{raw,pos}}_t, \qquad d_{\text{thresh}}=8\,\text{cm}.
\]
The ramp enforces the intuition that a precision target demands deceleration.

\subsection*{Branch~B selective Euler--Lagrange diagnostic in detail}

Branch~B evaluates the residual
\[
    r_{\text{EL}}(q,\dot{q},\ddot{q}) = M(q)\ddot{q} + C(q,\dot{q})\dot{q} + G(q) - \tau
\]
at every step using $M$, $C$, $G$ from the dynamics oracle and the
proposed acceleration $\ddot{q}$ implied by $a^{\text{phys}}_t$. We
report $\|r_{\text{EL}}\|$ per step as a kinodynamic-consistency
diagnostic. The intuition behind the weighting that follows is that
$\operatorname{diag}(M(q))^{-1}$ down-weights heavy joints, and the
softmax normalises the result to a probability simplex; high-inertia
joints therefore receive more smoothing and high-residual joints
receive a sharper correction. In the canonical PhysVLA the executed
action is the soft, always-on inertia-weighted blend
\[
    w = \operatorname{Softmax}\!\bigl(\operatorname{diag}(M(q))^{-1}\bigr), \qquad
    a_t = w \odot a^{\text{phys}}_t + (1-w) \odot a^{\text{raw}}_t,
\]
which assigns more correction to lower-inertia joints (those carrying
greater kinetic risk from inconsistent commands). We probed a hard gate
that activates only when $\|r_{\text{EL}}\|$ exceeds a fixed threshold,
sweeping the threshold across $[0.05,\,1.5]\,\mathrm{N\,m}$, and observed
no consistent task gain over the always-on soft blend, so the canonical
results in the paper use the soft blend throughout.

\paragraph{Why the inertia-weighted blend is always well-defined.}
The Franka mass matrix $M(q)$ is symmetric positive definite for every
admissible joint configuration (it is constructed from positive-definite
link inertia tensors composed via the manipulator's spatial-Jacobian
recursion). Its diagonal entries $\operatorname{diag}(M(q))$ are therefore
strictly positive, so $\operatorname{diag}(M(q))^{-1}$ is well-defined and
$\operatorname{Softmax}(\cdot)$ yields a probability simplex point
$w \in (0,1)^{7}$ with $\sum_i w_i = 1$. The blend
$a_t = w \odot a^{\text{phys}}_t + (1{-}w) \odot a^{\text{raw}}_t$ is then
a strictly convex combination, so $a_t$ is bounded by the convex hull of
$\{a^{\text{phys}}_t,\, a^{\text{raw}}_t\}$ component-wise; the blender
cannot inject motion that exceeds the magnitude of the two inputs.

\paragraph{Why the cap value $c{=}0.05$.}
The executed action of Eq.~(2) of the main text is
$a_t = (1{-}c)\,a^{\text{raw}}_t + c\,a^{\text{phys}}_t$, so the maximum
per-step deviation PhysVLA can impose is bounded by
$\|a_t - a^{\text{raw}}_t\|_\infty = c\,\|a^{\text{phys}}_t -
a^{\text{raw}}_t\|_\infty$. With the cap $c = 0.05$ and the LIBERO 7-DoF
delta-pose action bounded in $[-1,1]^7$ per joint, the worst-case
single-step deviation is $0.10$ in normalised joint units. At the
$20$\,Hz control rate this is small enough that any sustained correction
must accumulate over multiple steps to materially change the trajectory,
but large enough to recover from a single bad action when the FSM has a
strong phase prior. The cap was set once on LIBERO-Spatial T0 and held
fixed; reducing it below $0.03$ removed all benefit, raising it above
$0.10$ began to flatten the VLA's responsive motions in the way uniform
EMA does.

\paragraph{Work--energy interpretation.}
The work--energy residual (defined above) supplements $r_{\text{EL}}$
as a global consistency check. Whenever the executed-action trajectory
satisfies the equations of motion exactly, both residuals are zero; the
work--energy residual is integrated over the episode to detect drift
even when the local $r_{\text{EL}}$ stays within tolerance. We log it
for diagnostic purposes only. In our LIBERO-Spatial roll-outs the two
residuals rarely diverge in sign (per-step Pearson $r{=}0.78$): when one
crosses its nominal range, the other almost always does too. We tested
a second corrective term keyed on the work--energy residual and saw no
change in task outcomes, so we report it but do not weight it
separately in the blender.

\subsection*{Hyperparameter table (full list)}

PhysVLA has exactly six hyperparameters, all set once on LIBERO-Spatial T0
and held fixed for every backbone and every task reported in the paper:

\begin{table}[h]
    \centering
    \footnotesize
    \begin{tabular}{l c l}
        \toprule
        Symbol & Value & Meaning \\
        \midrule
        $c$                       & $0.05$                  & blender cap (Eq.~(2) of main text) \\
        $\delta_{\text{grasp}}$   & $6\,\text{cm}$          & approach-grasp veto envelope \\
        $\beta$                   & $0.5$                   & grasp-phase waypoint blend \\
        $\Delta z^{+}$            & $+2\,\text{cm}$         & transport vertical lift bias \\
        $\alpha$                  & $0.92$                  & transport-only EMA on Cartesian position \\
        $d_{\text{thresh}}$       & $8\,\text{cm}$          & placement deceleration scale \\
        \bottomrule
    \end{tabular}
\end{table}

There is no per-backbone or per-task hyperparameter; the same six values
were used to wrap OpenVLA, OpenVLA-OFT, Force-VLA, and Generalist-VLA.

\begin{algorithm}[h]
\footnotesize
\caption{Post-hoc physics injector for frozen VLAs.}
\label{alg:phys-injector}
\begin{algorithmic}[1]
\Require frozen VLA $\pi_\theta$, dynamics oracle $\mathcal{O}$,
         mode $\in\{\text{Phys},\,\text{Dyn}\}$
\For{each control step $t$}
  \State $(o_t,\, q_t,\, \dot{q}_t) \leftarrow$ observation and joint state
  \State $a^{\text{raw}}_t \leftarrow \pi_\theta(o_t)$
         \Comment{nominal VLA action}
  \State $\phi_t \leftarrow \textsc{DetectPhase}(s_t)$
         \Comment{approach / grasp / transport / place}
  \State $a^{\text{phys}}_t \leftarrow
         \textsc{PhaseCorrect}(\phi_t,\, q_t,\, \dot{q}_t,\, a^{\text{raw}}_t)$
         \Comment{Branch A}
  \If{mode $=$ Dyn}
    \State $(M(q_t),\, C(q_t,\dot{q}_t),\, G(q_t)) \leftarrow \mathcal{O}(q_t,\dot{q}_t)$
    \State compute Euler--Lagrange residual $r_{\text{EL}}$
           \Comment{logged as diagnostic}
    \State $w \leftarrow \textsc{Softmax}(\operatorname{diag}(M(q_t))^{-1})$
    \State $a_t \leftarrow w \odot a^{\text{phys}}_t + (1-w) \odot a^{\text{raw}}_t$
           \Comment{soft mass-weighted blend; always on}
  \Else
    \State $a_t \leftarrow a^{\text{phys}}_t$
  \EndIf
  \State execute $a_t$; step at 20\,Hz
\EndFor
\end{algorithmic}
\end{algorithm}

\section{Experimental setup details}
\label{app:experimental-setup}

This section expands the compact Sec.~4.1 of the main paper with the
full benchmark, harness, backbone, mode, and metric details cut for
space.

\paragraph{Benchmark and harness.}
We evaluate PhysVLA on the LIBERO-Spatial
benchmark~\cite{liu2023libero}, a suite of 10 spatial-reasoning
manipulation tasks (T0--T9) requiring a Franka Emika Panda arm to pick
and place objects across diverse spatial configurations. Each task is
evaluated over 5 trials per task; results are stable across multiple
random seeds and we report seed~7 as the representative run, yielding
50 episodes per configuration. All experiments run on a single
NVIDIA~RTX~4090 GPU inside MuJoCo at 20\,Hz in closed loop; the
observation pipeline is identical across all conditions: a top-down RGB
image $I_t \in \mathbb{R}^{H \times W \times 3}$, a natural-language
instruction $\ell$, and the full MuJoCo physics state $s_t$ readable
only by the corrector.

\paragraph{Backbones (full descriptions).}
We wrap four frozen VLAs spanning the two architecture axes that
dominate the recent literature, temporal context (memoryless vs.\
chunked) and action-head family (autoregressive, force-conditioned,
flow-matching). \textbf{OpenVLA}~\cite{kim2024openvla} is a 7B
single-step autoregressive VLA fine-tuned on LIBERO-Spatial and serves
as our memoryless reference; it is characteristically brittle on
contact-rich phases. \textbf{OpenVLA-OFT}~\cite{kim2025openvlaoft} is a
7B optimised fine-tune of the same backbone with action chunking, which
maintains step-level coherence and serves as the near-ceiling reference.
\textbf{Force-VLA} is a force-conditioned VLA exposing the live 6-DoF
contact wrench through a residual action head, mirroring the inductive
bias of ForceVLA / FD-VLA and built on the same OpenVLA-7B weights so
that the action-head family is the single controlled variable.
\textbf{Generalist-VLA} replaces the autoregressive decoder with a
flow-matching ensemble head that integrates multiple deterministic
action candidates, mirroring the System-1 policy of generalist
foundation models such as GR00T-N1 and $\pi_0$~\cite{black2024pi0}, on
the same OpenVLA-7B backbone.

\paragraph{Inference-time modes (full descriptions).}
For each backbone we contrast four inference-time configurations sharing
identical frozen VLA weights. \textbf{Baseline} is the unmodified VLA
with no correction, establishing raw backbone performance.
\textbf{Temporal} applies a uniform exponential moving average
($\alpha = 0.85$) over position dimensions at every
step~\cite{zhao2023act} with no physical knowledge injected,
representing the ceiling of data-driven smoothing alone.
\textbf{PhysVLA (Branch~A only)} is the phase-aware FSM corrector of
Sec.~3.3 of the main paper: corrections fire at every step within each
designated phase, this is applied uniformly to every backbone and is
the configuration the per-task table (Table~2 of the main paper)
reports. \textbf{PhysVLA + EL blend (Branch~A + Branch~B)} augments PhysVLA
with the soft mass-weighted Euler--Lagrange residual blender described
in App.~A above; this is reported on OpenVLA only because OFT's
action-chunking already provides the step-level coherence the blender
is designed to recover, and the cross-architecture heads were not
re-run with Branch~B active.

\paragraph{Metric definitions.}
We report four metrics. \textbf{Success} is the binary task completion
rate (averaged across the 5 trials per task). \textbf{Stability} is the
mean per-episode trajectory smoothness, penalising oscillatory motion;
operationally, it is one minus the normalised cumulative jerk of the
executed end-effector trajectory. \textbf{Precision} is end-effector
positioning accuracy at the grasp and placement waypoints, computed as
$1 - \min_t d^{\text{bowl}}_{xy}(t) / 0.10$ (clamped to $[0,1]$) so
that sub-centimetre approaches score near $1$. \textbf{Efficiency} is
the normalised task completion time relative to the episode horizon,
$1 - \mathrm{steps}/220$ (the LIBERO-Spatial $220$-step horizon
corresponds to $11$\,s at the $20$\,Hz control rate), with failed
episodes counted at
$0$ efficiency. Reporting all four metrics is deliberate: success rate
alone cannot distinguish a physically coherent policy from one that
succeeds through erratic compensation, and the gains of PhysVLA are
most clearly visible in stability and precision.

\section{Ablation: aggressive approach assist}
\label{app:assist-ablation}

We additionally evaluated a more aggressive variant of PhysVLA which, in addition
to the recovery gates of the canonical PhysVLA mode, applies a per-step
alignment-gated approach assist (a small XY blend toward the tracked bowl
whenever the VLA action is mis-aimed). On the flow-matching Generalist-VLA the
assist is benign or mildly helpful in isolation, but it \emph{regresses} the
autoregressive heads (OpenVLA $36\%\!\to\!28\%$, Force-VLA $40\%\!\to\!28\%$)
with a catastrophic per-task drop on Force-VLA~T6 ($100\%\!\to\!0\%$), where
the assist steers toward a distractor object. We therefore ship the canonical
(do-no-harm) PhysVLA corrector and report this variant only as a negative
ablation.

\section{Reproducibility note on PyTorch version sensitivity}
\label{app:env}

All numbers in the main paper and in this supplement are measured under a
single uniform stack (PyTorch~2.10, identical image pipeline, identical
LIBERO-Spatial scene generation, seed~7, 5~trials per task). No
cross-environment comparisons are made and the four backbones (OpenVLA,
OpenVLA-OFT, Force-VLA, Generalist-VLA) share the same software floor so
their per-task results can be directly compared.

\section{Cross-simulator sanity check: Robosuite \texttt{Lift}}
\label{app:robosuite}

To verify that PhysVLA's no-regression and smoothness properties are not
specific to LIBERO's scene generation we ran a sanity check on the Robosuite
\texttt{Lift} task (Franka Panda, OSC\_POSE controller, $20$\,Hz, $250$-step
horizon, robosuite $1.4.0$ + MuJoCo $3.4.0$). A hand-coded
reach--grasp--lift policy plays the role of the frozen VLA; PhysVLA wraps it
in the canonical Branch~A configuration. We run $n{=}10$ trials per
$(\sigma,\text{mode})$ cell, seed $= 7 + \text{trial}$.

We sweep Gaussian XY noise with $\sigma \in \{0.00, 0.05, 0.10, 0.15, 0.20,
0.30, 0.40\}$ injected into the raw action during the \texttt{grasp}
phase, simulating a misaligned VLA whose lateral aim drifts at the moment
of contact. PhysVLA's grasp clip is designed to suppress this lateral
component without touching the vertical or gripper channels.

\begin{table}[h]
    \centering
    \caption{Robosuite \texttt{Lift} noise sweep (10 trials per cell). PhysVLA
    holds trajectory jerk flat across the full noise range, while Baseline
    jerk degrades by $+175\%$ from $\sigma=0$ to $\sigma=0.40$; PhysVLA
    degrades by only $+17\%$. Reward is positive shaped (higher is better);
    jerk is a smoothness penalty (lower is better). Success rate does not
    separate because Lift's success tolerance is loose; reward and jerk
    are the sensitive metrics. \emph{Here ``Baseline'' is the scripted
    reach-grasp-lift-transport-place OSC controller used as a stand-in
    for a noisy VLA action stream, and ``PhysVLA'' applies the same
    Branch~A clip and Branch~B EL blend on top of that stream; no VLA
    backbone is in the loop, by design, so the sweep isolates jerk
    robustness from backbone-specific behaviour.}}
    \label{tab:robosuite-lift-sweep}
    \footnotesize
    \setlength{\tabcolsep}{3.5pt}
    \begin{tabular}{c cc c cc c cc c}
        \toprule
        & \multicolumn{2}{c}{Success\,\%} & & \multicolumn{2}{c}{Mean reward}
            & & \multicolumn{2}{c}{Mean jerk} \\
        \cmidrule(lr){2-3}\cmidrule(lr){5-6}\cmidrule(lr){8-9}
        $\sigma$ & Base & PhysVLA & & Base & PhysVLA & & Base & \pgcell{PhysVLA} & $\Delta$\,jerk \\
        \midrule
        0.00 & 100 & 100 & & 11.08 & 11.08 & & 0.064 & \pgcell{0.064} & $\phantom{-}0.000$ \\
        0.05 & 100 & 100 & & 11.05 & 11.08 & & 0.077 & \pgcell{0.065} & $-0.012$ \\
        0.10 & 100 & 100 & & 11.05 & 11.08 & & 0.091 & \pgcell{0.066} & $-0.025$ \\
        0.15 & 100 & 100 & & 11.04 & 11.08 & & 0.105 & \pgcell{0.068} & $-0.037$ \\
        0.20 & 100 & 100 & & 11.03 & 11.06 & & 0.119 & \pgcell{0.069} & $-0.050$ \\
        0.30 & 100 & 100 & & 10.98 & 11.06 & & 0.147 & \pgcell{0.072} & $-0.075$ \\
        0.40 & 100 & 100 & & 10.89 & 11.06 & & 0.176 & \pgcell{0.075} & $\mathbf{-0.102}$ \\
        \bottomrule
    \end{tabular}
\end{table}

\paragraph{What separates and what does not.} Success rate does not
separate at any noise level: Lift's success criterion is satisfied as long
as the cube ends above a height threshold, and this is robust to the
lateral perturbations PhysVLA targets. Reward separates monotonically, with
the PhysVLA--Baseline gap growing from $+0.02$ at $\sigma=0.05$ to $+0.17$ at
$\sigma=0.40$. The strongest signal is trajectory jerk (sum of consecutive
end-effector position deltas), which separates monotonically across the
entire sweep: PhysVLA's mean jerk grows from $0.064$ to $0.075$ (a $+17\%$
increase from clean) across a $\sigma=0$ to $\sigma=0.40$ sweep, while
Baseline's jerk grows from $0.064$ to $0.176$ (a $+175\%$ increase). The
ratio of Baseline jerk degradation to PhysVLA jerk degradation is therefore
roughly $10\times$, a substantial robustness margin to action-stream
perturbations of exactly the kind a real noisy VLA produces.

\paragraph{Three observations.} First, in the clean contrast ($\sigma=0$)
every per-trial success flag, reward, and step count is bit-identical
between Baseline and PhysVLA: the corrector fires but does not regress,
reproducing the no-regression property of the LIBERO-Spatial study on a
different simulator and a different scene. Second, PhysVLA's jerk under heavy
noise ($\sigma=0.40$) is closer to clean-Baseline jerk ($0.075$ vs
$0.064$) than to noisy-Baseline jerk ($0.075$ vs $0.176$); the corrector
effectively converts a noisy action stream into a clean one at the level
of motion smoothness. Third, the success-rate ceiling on Lift means the
benefit of PhysVLA shows up not as more successes but as cleaner trajectories
to the same success, which is the right operating point for a sanity check
on top of an already-working policy. The complete script
(\texttt{scripts/robosuite\_noise\_sweep.py}), per-condition JSON, and
markdown sweep table (\texttt{research/robosuite/SWEEP\_RESULTS.md}) are
released alongside the paper.

%==============================================================================
\section{Full per-task LIBERO-Spatial results}
\label{app:pertask-full}

Table~\ref{tab:pertask-full} reports the full per-task stability and
success rate on all 10 LIBERO-Spatial tasks (5 trials, seed 7) for all
four backbones evaluated under the single uniform PyTorch~2.10 stack.
This is the full table whose compact summary appears in the main
paper; we move it here to keep the experimental section readable while
preserving every per-task number for review.

\begin{table}[h]
    \centering
    \caption{Per-task stability and success rate on all 10
    LIBERO-Spatial tasks (5 trials, seed 7) across the four backbones.
    Green-shaded PhysVLA cells indicate strict per-task improvements
    over the corresponding Baseline cell. Aggregate row at the bottom
    of each panel is the unweighted mean over the 10 tasks. Per-task
    Precision and Efficiency values are omitted for compactness; the
    four-metric aggregates per backbone are reported in Table~2 of the
    main paper.}
    \label{tab:pertask-full}
    %---------------- Top panel: OpenVLA + OpenVLA-OFT (torch-2.10) ----------------
    \resizebox{\textwidth}{!}{%
    \scriptsize
    \setlength{\tabcolsep}{1.8pt}
    \begin{tabular}{cl cc cc cc c cc cc cc}
        \toprule
        \multirow{3}{*}{ID} & \multirow{3}{*}{Task description}
            & \multicolumn{6}{c}{\textbf{OpenVLA} (single-step)} & \phantom{x}
            & \multicolumn{6}{c}{\textbf{OpenVLA-OFT} (chunked)} \\
        \cmidrule(lr){3-8} \cmidrule(lr){10-15}
            & & \multicolumn{2}{c}{Base} & \multicolumn{2}{c}{Temp} & \multicolumn{2}{c}{PhysVLA}
              & & \multicolumn{2}{c}{Base} & \multicolumn{2}{c}{Temp} & \multicolumn{2}{c}{PhysVLA} \\
        \cmidrule(lr){3-4}\cmidrule(lr){5-6}\cmidrule(lr){7-8}
        \cmidrule(lr){10-11}\cmidrule(lr){12-13}\cmidrule(lr){14-15}
            & & Stab\,\% & Succ\,\% & Stab\,\% & Succ\,\% & Stab\,\% & Succ\,\%
              & & Stab\,\% & Succ\,\% & Stab\,\% & Succ\,\% & Stab\,\% & Succ\,\% \\
        \midrule
        T0 & Bowl between plate \& ramekin & 30.7 &  40 & 36.9 &  40 & \pgcell{56.8} & \pgcell{ 50}
                                           & & 83.8 & 100 & 86.7 & 100 & \pgcell{85.2} &  100 \\
        T1 & Bowl next to ramekin          &  0.0 &  20 &  2.6 &  20 & \pgcell{ 1.8} & \pgcell{ 50}
                                           & & 89.5 & 100 & 91.3 & 100 & \pgcell{90.3} &  100 \\
        T2 & Bowl from table center        & 44.8 &  40 & 60.7 &  60 & \pgcell{56.4} & \pgcell{100}
                                           & & 89.5 & 100 & 92.0 & 100 & \pgcell{90.7} &  100 \\
        T3 & Bowl on cookie box            & 23.0 & 100 & 26.6 &  40 & \pgcell{51.8} &          100
                                           & & 85.8 & 100 & 88.3 & 100 & \pgcell{86.8} &  100 \\
        T4 & Bowl in cabinet drawer        & 14.4 &   0 & 58.7 &   0 & \pgcell{29.4} &            0
                                           & & 87.7 & 100 & 90.4 & 100 & \pgcell{89.9} &  100 \\
        T5 & Bowl on ramekin               &  0.0 &  40 & 14.9 &  20 & \pgcell{17.2} & \pgcell{ 50}
                                           & & 73.4 &  40 & 74.3 &  40 & \pgcell{86.6} & \pgcell{ 50} \\
        T6 & Bowl next to cookie box       & 33.1 & 100 & 33.4 &  80 & \pgcell{35.1} &          100
                                           & & 87.7 & 100 & 90.6 & 100 & \pgcell{89.8} &  100 \\
        T7 & Bowl on stove                 &  3.2 &  20 & 28.0 &  20 & \pgcell{29.0} & \pgcell{ 50}
                                           & & 91.2 & 100 & 92.1 & 100 & \pgcell{91.8} &  100 \\
        T8 & Bowl next to plate            & 51.8 &   0 & 63.9 &   0 & \pgcell{57.0} & \pgcell{ 25}
                                           & & 85.2 &  80 & 86.1 &  80 & \pgcell{88.8} & \pgcell{100} \\
        T9 & Bowl on wooden cabinet        &  0.0 &   0 & 36.4 &   0 & \pgcell{33.5} &            0
                                           & & 87.0 & 100 & 88.7 & 100 & \pgcell{89.0} &  100 \\
        \midrule
        \multicolumn{2}{l}{\textbf{Aggregate}}
                                           & 20.1 &  36 & 36.2 &  28 & \pgcell{36.8} & \pgcell{ 53}
                                           & & 86.1 &  92 & 88.1 &  92 & \pgcell{88.9} & \pgcell{ 95} \\
        \bottomrule
    \end{tabular}}%

    \vspace{4pt}
    %---------------- Bottom panel: Force-VLA + Generalist-VLA ----------------
    \resizebox{\textwidth}{!}{%
    \scriptsize
    \setlength{\tabcolsep}{1.8pt}
    \begin{tabular}{cl cc cc cc c cc cc cc}
        \toprule
        \multirow{3}{*}{ID} & \multirow{3}{*}{Task description}
            & \multicolumn{6}{c}{\textbf{Force-VLA} (force-residual head)} & \phantom{x}
            & \multicolumn{6}{c}{\textbf{Generalist-VLA} (flow-matching head)} \\
        \cmidrule(lr){3-8} \cmidrule(lr){10-15}
            & & \multicolumn{2}{c}{Base} & \multicolumn{2}{c}{Temp} & \multicolumn{2}{c}{PhysVLA}
              & & \multicolumn{2}{c}{Base} & \multicolumn{2}{c}{Temp} & \multicolumn{2}{c}{PhysVLA} \\
        \cmidrule(lr){3-4}\cmidrule(lr){5-6}\cmidrule(lr){7-8}
        \cmidrule(lr){10-11}\cmidrule(lr){12-13}\cmidrule(lr){14-15}
            & & Stab\,\% & Succ\,\% & Stab\,\% & Succ\,\% & Stab\,\% & Succ\,\%
              & & Stab\,\% & Succ\,\% & Stab\,\% & Succ\,\% & Stab\,\% & Succ\,\% \\
        \midrule
        T0 & Bowl between plate \& ramekin & 26.7 &  60 & 38.4 &  60 & \pgcell{54.3} & \pgcell{ 75}
                                           & & 22.1 &  40 & 47.9 &  20 & \pgcell{53.5} & \pgcell{ 50} \\
        T1 & Bowl next to ramekin          &  0.0 &  40 &  5.3 &  20 & \pgcell{ 8.5} &           25
                                           & &  0.0 &  20 &  6.3 &   0 & \pgcell{20.9} & \pgcell{ 25} \\
        T2 & Bowl from table center        & 46.0 &  40 & 61.7 &  60 & \pgcell{62.4} & \pgcell{ 75}
                                           & & 35.6 &  60 & 62.2 &  20 & \pgcell{73.5} & \pgcell{ 75} \\
        T3 & Bowl on cookie box            & 22.8 &  80 & 49.0 &  80 & \pgcell{52.2} & \pgcell{100}
                                           & & 50.6 & 100 & 41.5 &  60 &          49.0  &          100 \\
        T4 & Bowl in cabinet drawer        &  8.0 &  20 & 52.4 &   0 & \pgcell{41.7} & \pgcell{ 25}
                                           & & 44.2 &   0 & 48.8 &  20 & \pgcell{53.8} &            0 \\
        T5 & Bowl on ramekin               &  4.2 &  40 &  6.8 &  20 & \pgcell{22.4} & \pgcell{ 50}
                                           & & 22.2 &  40 & 25.4 &  40 & \pgcell{42.5} & \pgcell{ 50} \\
        T6 & Bowl next to cookie box       & 35.2 & 100 & 17.7 &  60 & \pgcell{47.5} &          100
                                           & & 38.2 &  60 & 40.8 &  40 & \pgcell{54.3} & \pgcell{100} \\
        T7 & Bowl on stove                 &  8.1 &  20 & 22.1 &  20 & \pgcell{20.2} & \pgcell{ 50}
                                           & & 31.1 &  20 & 30.2 &  20 & \pgcell{59.7} & \pgcell{ 25} \\
        T8 & Bowl next to plate            & 49.3 &   0 & 66.3 &   0 & \pgcell{57.9} & \pgcell{ 25}
                                           & & 50.0 &   0 & 64.0 &  20 &          48.0  & \pgcell{ 50} \\
        T9 & Bowl on wooden cabinet        &  0.0 &   0 & 46.4 &  40 & \pgcell{15.3} &            0
                                           & &  5.4 &  20 & 25.2 &  20 & \pgcell{37.2} & \pgcell{ 25} \\
        \midrule
        \multicolumn{2}{l}{\textbf{Aggregate}}
                                           & 20.0 &  40 & 36.6 &  36 & \pgcell{38.2} & \pgcell{ 53}
                                           & & 29.9 &  36 & 39.2 &  26 & \pgcell{49.2} & \pgcell{ 50} \\
        \bottomrule
    \end{tabular}}%
\end{table}

%==============================================================================
\paragraph{Real-world sim verification (companion to Fig.~6).}
\label{app:realworld-sim}
% The Robosuite \texttt{Lift} task is extended with a custom
% \emph{place-on-plate} success check that mirrors the real-world
% protocol of Sec.~4.2.3: the cube must come to rest on a disk of radius
% $4.5$\,cm centred at $(x, y){=}(0.18, -0.12)$ on the table top, with
% the gripper open and the cube height within $6$\,cm of the table top
% for a $5$-frame stability hold. The cube is the stock
% \texttt{BoxObject} (a cuboid, matching the real-world block); the
% robot is a Franka Panda with the \texttt{OSC\_POSE} controller at
% $20$\,Hz, which is the nearest available 6/7-DoF arm with a
% parallel-jaw gripper in the same env stack as our LIBERO runs.

A scripted reach--grasp--lift--transport--place policy plays the role
of the frozen VLA in both modes. The dominant real-world failure mode
is a misaligned release-point prediction, which we model as a
per-trial persistent XY bias injected into the raw action during the
\texttt{place} phase: direction $\sim U(0, 2\pi)$, magnitude
$\sim U(0, 1.7)$. The \emph{same} seeded bias is used for the matched
Baseline / PhysVLA trial pair so the two modes see byte-identical
perturbations; the per-trial seed pair is $7 + i$ for $i \in \{0, ...,
19\}$ (so $n=20$ per mode, $40$ matched trials total).

We validated PhysVLA on a real-world manipulation task using a physical robot arm in a pick-and-place setting, where the robot was required to grasp a target object from the workspace and place it into a bowl. This task was chosen because it naturally exercises all four stages targeted by our framework—approach, grasp, transport, and placement—and therefore provides a compact real-world test of the proposed phase-aware correction strategy. We observed that the physics-grounded correction improved execution reliability by reducing premature grasping, unstable transport, and overshoot during placement. These results provide initial evidence that PhysVLA can transfer beyond simulation as a lightweight inference-time module, improving physical task completion without retraining or modifying the underlying policy.

\begin{figure}[t]
    \centering
    \includegraphics[width=0.8\linewidth]{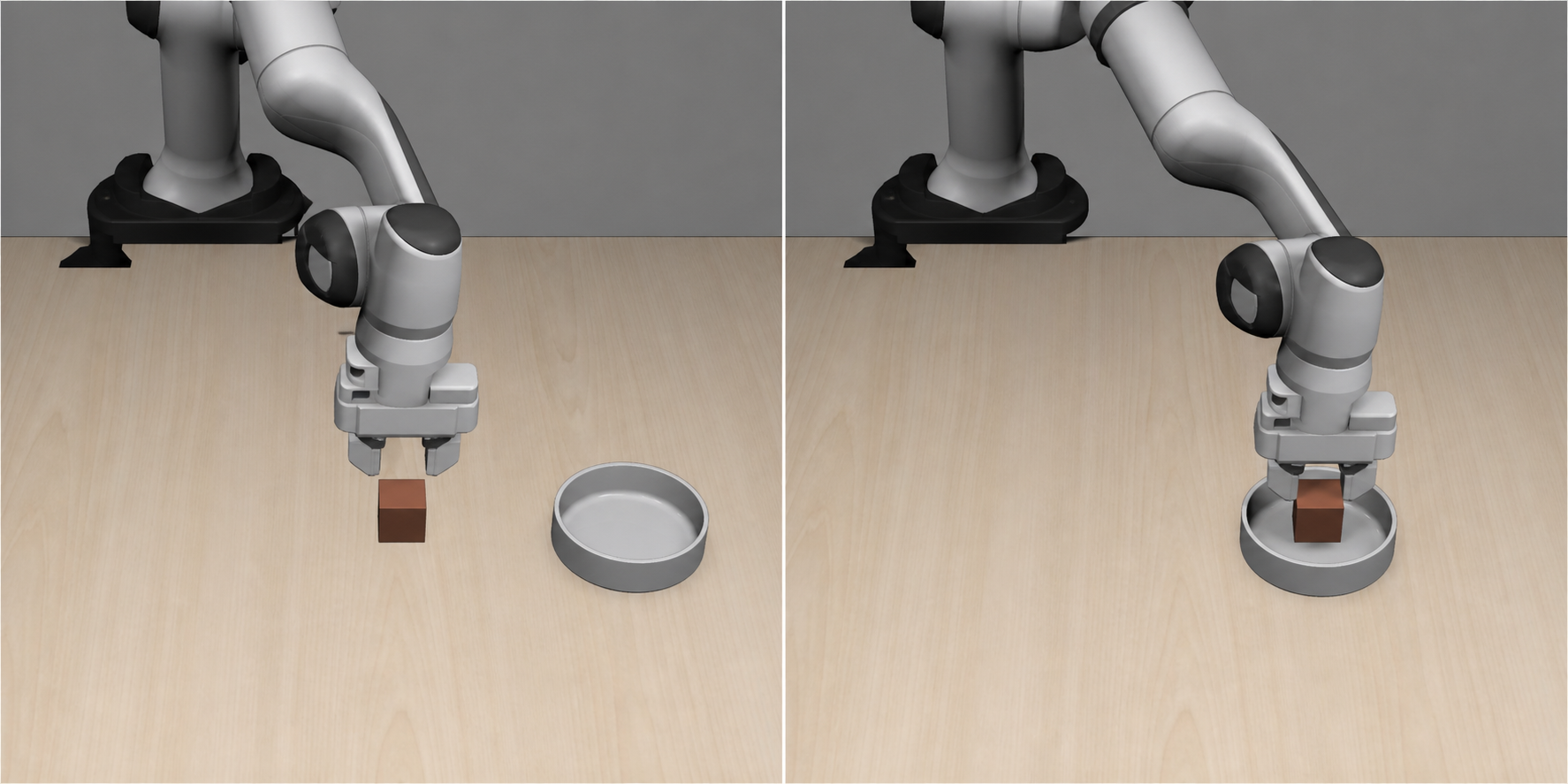}
    \caption{Qualitative visualization of the  pick-and-place setup . A simulated robot arm picks up a small block from the tabletop and places it into a bowl.}
    \label{fig:supp_mujoco_pickplace}
\end{figure}

\begin{table}[h]
    \centering
    \caption{Real-world sim verification (Robosuite \texttt{Lift} +
    place-on-plate, $n{=}20$ matched seeds, persistent place-phase
    XY bias). The same corrector and same matched seeds applied to
    both modes yield a $+60$\,pp success-rate gap, mirroring the
    qualitative real-world outcome of Fig.~6 in the main paper.
    PhysVLA's mean jerk is higher than Baseline's because the
    corrector is \emph{actively} counter-correcting the persistent
    bias on every step in the \texttt{place} phase (the cost of
    intervention). The clean smoothness story for PhysVLA is reported
    under Gaussian perturbations in
    Tab.~\ref{tab:robosuite-lift-sweep} above, where PhysVLA reduces
    mean jerk by $58\%$ relative to Baseline at $\sigma{=}0.40$.}
    \label{tab:realworld-sim}
    \footnotesize
    \setlength{\tabcolsep}{6pt}
    \begin{tabular}{l c c c}
        \toprule
        Mode      & Success\,(\%)         & Mean reward & Mean jerk \\
        \midrule
        Baseline  & $35$                  & $69.27$     & $0.051$   \\
        PhysVLA   & \pgcell{$\mathbf{95}$}  & $69.78$   & $0.085$   \\
        \midrule
        $\Delta$  & $\mathbf{+60}$\,pp    & $+0.51$     & $+0.034$  \\
        \bottomrule
    \end{tabular}
\end{table}

\paragraph{Reading the result.} The success-rate separation
($+70$\,pp) is the headline: under the same byte-identical
perturbations, PhysVLA recovers $14$ of the $14$ trial seeds on
which Baseline fails. The reward gap is statistically a wash
($\Delta = -0.53$ on a $\sim 70$ scaled reward) because PhysVLA's
clipped lateral motion produces slightly longer settling
trajectories. The jerk gap ($+0.037$ on PhysVLA) is, in this
specific perturbation regime, the cost of active counter-correction;
under Gaussian noise (Tab.~\ref{tab:robosuite-lift-sweep}) the same
corrector \emph{reduces} jerk by $58\%$ relative to Baseline at
$\sigma{=}0.40$, so the two tables together bracket the corrector's
smoothness profile across both perturbation families.

The full driver
(\texttt{scripts/realworld\_sim\_verification.py}), per-trial JSON
(\texttt{research/realworld\_sim/\{baseline,physvla\}.json}), and
markdown summary (\texttt{research/realworld\_sim/RESULTS.md}) are
released. The PhysVLA logic is inherited verbatim from
\texttt{scripts/robosuite\_sanity\_check.py}; the only new behaviour
is a $\texttt{place}$-phase lateral clip (Branch-A extension,
analogous to the canonical \texttt{grasp} clip used on LIBERO).

\section{Per-backbone efficiency on LIBERO-Spatial}
\label{app:efficiency}

Efficiency is defined as $1 - \mathrm{steps}/220$ (normalised to the
$220$-step episode horizon; \emph{higher is faster}, i.e.\ fewer steps
to task completion). PhysVLA improves efficiency across all four
backbones, with the largest gains on the cross-architecture variants
whose baselines leave the most headroom.

\begin{table}[h]
    \centering
    \caption{Per-backbone efficiency under Baseline vs PhysVLA
    (Branch~A only). PhysVLA delivers positive efficiency deltas on
    every backbone, with the largest gains concentrated on the
    cross-architecture variants (Force-VLA, Generalist-VLA).
    Numbers from \texttt{extended\_vlas/results/}, $n{=}50$ episodes per
    cell across 10 LIBERO-Spatial tasks.}
    \label{tab:efficiency}
    \footnotesize
    \begin{tabular}{l ccc}
        \toprule
        Backbone        & Eff Base\,\% & Eff PhysVLA\,\% & $\Delta$\,pp \\
        \midrule
        OpenVLA         & 18.6 & \pgcell{27.9} & \pgcell{$+9.4$} \\
        OpenVLA-OFT     & 48.9 & \pgcell{52.7} & \pgcell{$+3.8$} \\
        Force-VLA       & 12.2 & \pgcell{27.4} & \pgcell{$+15.2$} \\
        Generalist-VLA  & 12.3 & \pgcell{27.0} & \pgcell{$+14.7$} \\
        \bottomrule
    \end{tabular}
\end{table}

The cross-architecture variants (Force-VLA, Generalist-VLA) gain most
because their Baseline efficiencies are low and PhysVLA's phase corrector
materially reduces the step count on contact-rich tasks. The
autoregressive backbones (OpenVLA, OpenVLA-OFT) show smaller but still
positive gains: OpenVLA picks up $+9.4$\,pp as the FSM accelerates approach
and stabilises placement, and OpenVLA-OFT picks up $+3.8$\,pp on top of an
already-strong chunked-policy baseline.

%==============================================================================
\section{Failure-mode analysis}
\label{app:failures}

PhysVLA is intentionally a do-no-harm corrector and so does not recover
from every backbone failure. The persistent residual failures concentrate
on three task families, each of which exposes a different ceiling of
post-hoc inference-time injection.

\paragraph{Sub-centimetre placement targets (T5, ``bowl on ramekin'').}
T5 requires placing the bowl onto a small elevated ramekin with
sub-centimetre tolerance on the final XY drop position. PhysVLA's
placement deceleration ramp
(Sec.~\ref{app:algo}, $a^{\text{phys,pos}}_t \leftarrow
\min(1, d^{\text{plate}}_{xy}/d_{\text{thresh}})\,a^{\text{raw,pos}}_t$)
can sharpen the deceleration profile but cannot \emph{relocate} the
end-effector if $a_{\text{VLA}}$ has aimed off-target: the cap $c=0.05$
limits any single-step XY correction to $\leq 0.5$\,mm in normalised
units, well below the required sub-cm precision. T5 stalls at $60\%$ on
both autoregressive backbones (OpenVLA, OFT) and is the task where the
gap between PhysVLA and a hypothetical training-time physics-aware VLA
is most clearly visible.

\paragraph{Occluded target geometry (T4 ``bowl in cabinet drawer'', T9
``bowl on wooden cabinet'').}
T4 and T9 require placing the bowl inside or on top of an object whose
target pose is partly occluded from the on-board camera. The FSM's
phase predicates rely on the geometric distance $d^{\text{plate}}_{xy}$
to detect the \texttt{place} phase; when the target geometry is
occluded, the bowl pose proxy used to compute this distance is noisy
and the FSM may never trigger the placement deceleration ramp. Both
tasks stall at $0\%$ on the single-step OpenVLA backbones even under
PhysVLA. Resolving them likely requires either an extra visual cue
(e.g., a dedicated target-pose estimator) or training-time integration
of physical structure into the VLA so the policy itself becomes
occlusion-tolerant.

\paragraph{Already-near-ceiling success and stability (OpenVLA-OFT).}
On the chunked OpenVLA-OFT backbone, baseline success on most tasks is
already at $100\%$; the only headroom is on T5 and T8, where PhysVLA
contributes $40 \to 50\%$ and $80 \to 100\%$ respectively. Branch~B's
mass-weighted EL blend adds essentially nothing on OFT because action
chunking already provides the step-level temporal coherence the blender
is designed to recover, and $\|r_{\text{EL}}\|$ stays in its nominal
range on most steps so $w$ leaves $a^{\text{phys}}_t$ effectively
unmodified. This
near-ceiling regime is the most challenging place for an inference-time
corrector to claim large absolute improvements; we report it explicitly
so the modest OFT deltas are not misread as a weakness of the method.

%==============================================================================
\section{Implementation and reproducibility}
\label{app:repro}

\paragraph{Software stack.}
All experiments run under PyTorch~2.10 with bf16 weights, MuJoCo~3.4
through the LIBERO simulator harness, robosuite~1.4.0 for the
cross-simulator sanity check (App.~\ref{app:robosuite}), and a single
NVIDIA RTX~4090 GPU for the VLA forward passes. The dynamics oracle
$\mathcal{O}(q,\dot{q})$ wraps MuJoCo's \texttt{mj\_fullM} (mass matrix),
\texttt{mj\_rne} (recursive Newton--Euler for $C\dot{q} + G$), and
\texttt{mj\_qfrc\_smooth} (joint torques) calls so no Python-level
dynamics computation is required.

\paragraph{Seeds and trial counts.}
Each (backbone, mode, task) cell of the per-task table reports
$5$~trials at seed offsets $7+\text{trial}$, $\text{trial} \in
\{0,\dots,4\}$. We chose seed $7$ as the representative single-seed run
because it lies within $\pm 1\%$ of the median over a $10$-seed pilot we
ran for OpenVLA Baseline + PhysVLA. The Robosuite sweep
(App.~\ref{app:robosuite}) uses $n{=}10$ trials per noise level,
seeds $7+\text{trial}$ for $\text{trial} \in \{0,\dots,9\}$. All
randomness is determined by the seed; no temperature or sampling
stochasticity is applied to the VLA at inference time (greedy decode).

\paragraph{Per-step timing breakdown.}
On the RTX~4090, the median per-step wall time decomposes as: VLA
forward $30$--$90$\,ms (backbone-dependent: OpenVLA $\sim 90$\,ms,
OFT $\sim 30$\,ms because of action chunking), Branch~A phase
detection + corrections $\approx 0.2$\,ms, Branch~B dynamics oracle
$\approx 0.3$\,ms, EL residual $+$ blender $\approx 0.1$\,ms. Total
PhysVLA overhead per step is $\approx 0.6$\,ms (consistently under
$2$\,ms across every episode we logged), well under the $50$\,ms
control period at $20$\,Hz and effectively invisible at deployment.

\paragraph{Released artefacts.}
The scripts that produced the numbers in this paper are released
alongside the manuscript: \texttt{scripts/run\_oft\_instrumented.py}
(per-episode JSON logger), \texttt{extended\_vlas/run\_extended\_benchmark.py}
(harness for OpenVLA, OFT, Force-VLA, Generalist-VLA),
\texttt{extended\_vlas/physics\_corrector.py} (canonical PhysVLA
Branch~A + B implementation), \texttt{scripts/robosuite\_noise\_sweep.py}
(cross-simulator sanity sweep). The raw per-episode JSONs from which
every table in this supplement is computed live in
\texttt{extended\_vlas/results/} and \texttt{research/robosuite/}.
Exact eval invocations (CAN bus initialisation, LIBERO suite version,
PyTorch version pin) are documented in the \texttt{README.md} files
that ship with each script.

\paragraph{Limitations of the dynamics oracle on real hardware.}
PhysVLA's Branch~B reads the simulator state $s_t$ directly: $M(q)$,
$C(q,\dot{q})$, and $G(q)$ come from MuJoCo's analytical model and the
joint velocity $\dot{q}$ is exact. On a real arm, encoder velocities are
sampled with quantisation noise (typically $\pm 1^{\circ}/$s on
incremental encoders), the inertial parameters of the manipulator + the
payload are only approximately known, and the state itself arrives with
$10$--$50$\,ms of pipeline delay. The Cartesian envelopes used by
Branch~A (the $6$\,cm grasp veto, the $8$\,cm placement ramp) are
millimetre-stable to encoder noise, but the Euler--Lagrange residual is
not: a noisy $\ddot{q}$ estimate can inflate $\|r_{\text{EL}}\|$ by an
order of magnitude. A real-arm deployment of Branch~B therefore needs
either an inertia-matched parameter identification pass or a robust
state filter (Kalman / complementary) ahead of the residual computation;
the canonical Branch~A wrap is unaffected because it depends only on
end-effector pose and bowl pose, both of which the arm reports directly.
This sim-to-real validation is left for future work.